%% file: main.tex
\documentclass{davian}

\input{preambles/libs}



\shortauthor{Lee et al.}
\runningtitle{Denoised Variance-Based Pruning with Optimal Brain Bias Compensation}

\date{\today}   % or e.g. \date{Preprint. \today} / \date{May 2026}

\title{Denoised Variance-Based Pruning with \\Optimal Brain Bias Compensation}
\authors{%
  Geon Tack Lee\affmark{1,2}\quad
  Jaegul Choo\affmark{1,\dag}\quad
  Kang Eun Jeon\affmark{1,\dag}%
}
\affiliations{%
  \affmark{1}DAVIAN Lab, KAIST AI \qquad
  \affmark{2}New York University Abu Dhabi%
}
\contributions{%
  \affmark{\dag}Corresponding authors.%
}
\correspondence{Email: \href{mailto:gtl256@nyu.edu}{gtl256@nyu.edu}, \href{mailto:jchoo@kaist.ac.kr}{jchoo@kaist.ac.kr}, \href{mailto:kejeon@kaist.ac.kr}{kejeon@kaist.ac.kr}}

\begin{document}
\maketitle

\input{sections/00_abstract}
\input{sections/01_introduction}
\input{sections/02_background_related_works}
\input{sections/03_method}
\input{sections/04_experiments_test_results}
\input{sections/05_conclusion}
%% ---- bibliography ----
\bibliographystyle{plainnat}
\bibliography{references}

%% ---- appendix (after the bibliography) ----
\input{sections/A_appendix}
\end{document}

%% file: preambles/libs.tex
\usepackage{amsfonts}          % extra math alphabets (\mathbb etc.)
\usepackage{dsfont}            % \mathds{1} indicator function
\usepackage{siunitx}           % \num{}, \SI{}{} — number & unit formatting
\usepackage{algorithm}         % float wrapper for pseudocode
\usepackage{ragged2e}          % \RaggedRight, \Centering, ...
\usepackage{subcaption}        % subfigures / subtables
\usepackage{wrapfig}           % text-wrapped figures and tables

\usepackage{tikz}              % used by the placeholder figure
\usepackage{lipsum}            % filler text

\graphicspath{{figs/}{tbls/}{assets/}}
\DeclareGraphicsExtensions{.pdf,.png,.jpeg,.jpg}

%% file: sections/00_abstract.tex
\begin{abstract}
Vision Transformers (ViTs) achieve state-of-the-art performance but carry massive computational overhead that restricts edge deployment.  
Although structural pruning has emerged as a key strategy to reduce these costs, existing methods often suffer from severe accuracy degradation or require expensive retraining. 
Recently, Variance-Based Pruning (VBP) introduced a promising paradigm by selecting neurons based on activation variance; however, it remains limited by statistical noise in finite-sample activation covariance and reliance on bias-only updates that cannot fully account for structural reconstruction error. 
To address these limitations, we introduce Denoised Variance-Based Pruning with Optimal Brain Bias Compensation (DVBP + OB$^2$C). 
We leverage random matrix theory to filter noise from the activation covariance spectrum for robust neuron selection and mathematically prove that integrating mean-shift compensation into the Optimal Brain Compression objective reduces the layer-wise Hessian exactly to the activation covariance matrix. 
This enables an optimal, closed-form update of the remaining weights using the same statistics gathered for selection. 
Extensive experiments on DeiT, Swin, and ConvNeXt architectures demonstrate that DVBP + OB$^2$C achieves state-of-the-art training-free performance; at 50\% MLP pruning, it retains over 90\% of the original Top-1 accuracy on Small and Base variants, outperforming VBP by up to 29.46\% (ConvNeXt-T) and 7.33\% (Swin-S). The code is available at: \url{https://github.com/geontackee/DVBP_OB2C}.
\keywords{Vision Transformers, Structured Pruning, Variance-Based Pruning}
\end{abstract}

%% file: sections/01_introduction.tex
\section{Introduction}
\label{sec:intro}
Since the introduction of the Vision Transformer (ViT)~\cite{Dosovitskiy2021ViT}, transformer-based architectures have consistently defined the state-of-the-art in computer vision, frequently outperforming traditional Convolutional Neural Networks (CNNs)~\cite{He2016ResNet}. However, this empirical success comes at a steep computational price. 
Standard ViTs demand substantial memory and computational resources for inference~\cite{Touvron2021DeiT}, effectively precluding their deployment on resource-constrained edge devices. 
For instance, ViT-B/16 requires 86 million parameters~\cite{Dosovitskiy2021ViT}, contrasting sharply with the 7.5 million of MobileNetV3~\cite{Howard2019MobileNetV3}. 
As vision models are increasingly deployed on-device for latency-critical applications, such as autonomous driving, robotics, and unmanned aerial systems, achieving smaller model footprints and higher inference throughput has become an indispensable research objective.

While pruning has emerged as a principled strategy to bridge this efficiency gap~\cite{Han2015Mag, Hoefler2021Sparsity}, current paradigms present a fundamental trade-off between theoretical sparsity and practical acceleration. 
Unstructured methods~\cite{Han2015Mag, Lee2019SNIP} achieve high compression ratios by zeroing individual weights. However, these methods produce irregular sparsity patterns that demand specialized hardware accelerators~\cite{Emani2023Accelerators} or custom CUDA kernels to induce practical latency reductions~\cite{Gale2019Sparsity}. This severely limits their scalability and real-world adoption. On the other hand, structured pruning~\cite{Li2017FilterPrune, Yu2023XPruner} physically removes entire neurons or filters, hence directly reducing model size and inference time on commodity hardware.
However, existing structured methods predominantly follow the magnitude-based pruning paradigm~\cite{Han2015Mag, Li2017FilterPrune}, which ranks parameters by the absolute magnitude of their weights under the assumption that smaller weights contribute less to the network. While this heuristic is effective for zeroing individual connections, removing entire neurons or filters based on aggregate magnitude discards complex feature interactions and leads to severely degraded accuracy~\cite{Luo2017Entropy}. 

In a fundamentally different direction, \textit{Variance-Based Pruning} (VBP)~\cite{Berisha2025VBP} has demonstrated that this trade-off need not be inevitable. 
Rather than relying on weight magnitudes, VBP introduces an activation variance criterion that identifies neurons whose activations exhibit the lowest variance across a calibration set. 
The key insight is that low-variance neurons behave approximately as constants, meaning their contribution to the downstream layer can be effectively absorbed into the bias term through a simple mean-shift compensation. 
This allows VBP to structurally remove neurons while immediately recovering a portion of the lost performance, substantially outperforming magnitude-based approaches. 
However, VBP still suffers from significant accuracy degradation at higher pruning ratios, leaving considerable room for improvement.

In this work, we identify two critical limitations of VBP. First, the activation covariance matrix, from which variance scores are derived, is corrupted by statistical noise. This noise obscures the true importance ranking of neurons, leading to increasingly suboptimal pruning decisions at aggressive compression ratios (\eg $\geq$50\%). 
Second, VBP compensates for pruned neurons solely through mean-shift compensation, leaving the remaining weight matrices entirely unmodified. 
We observe that mean-shift compensation cannot fully account for the reconstruction error introduced by removing entire neurons, and a principled compensation through weight update is necessary to close this gap.

To address these limitations, we propose Denoised Variance-Based Pruning with Optimal Brain Bias Compensation (DVBP + $\text{OB}^2\text{C}$). Our core contributions are:
\begin{itemize}
    \item \textbf{Denoised Variance Metric:} We leverage random matrix theory, specifically the Marchenko-Pastur distribution, to separate signal eigenvalues from noise in the activation covariance spectrum. The resulting denoised variance scores yield a substantially more reliable neuron importance ranking, particularly at high pruning ratios.
    \item \textbf{$\text{OB}^2\text{C}$ Framework:} We incorporate mean-shift compensation directly into the seminal Optimal Brain Compression (OBC)~\cite{Frantar2022OBC} reconstruction objective. This integration yields a remarkably elegant result: the layer-wise Hessian reduces exactly to the activation covariance matrix. Consequently, the same covariance matrix simultaneously serves as the basis for both neuron selection (via denoised variance) and optimal multi-weight recovery, unifying the two stages of our pipeline into a single, coherent statistical framework.
\end{itemize}
Extensive experiments on DeiT~\cite{Touvron2021DeiT}, Swin~\cite{Liu2021Swin}, and ConvNeXt~\cite{Liu2022ConvNeXt} architectures across Tiny, Small, and Base scales demonstrate that our method consistently outperforms VBP by a widening margin as the pruning ratio increases. At a 50\% structural reduction, DVBP + $\text{OB}^2\text{C}$ retains approximately 80\% Top-1 accuracy on Tiny-scale models and exceeds 90\% on Small and Base variants, surpassing VBP by up to 7.33\% on Swin-S and 29.46\% on ConvNeXt-T.

%% file: sections/02_background_related_works.tex
\section{Background and Related Works}
\label{sec:background_related_works}
\subsection{Model Compression and Pruning}
Model compression relies primarily on quantization~\cite{Frantar2023GPTQ, Ashkboos2024QuaRot} and pruning~\cite{Hoefler2021Sparsity, Lee2025Elsa}.
Unlike quantization, which reduces bit-width but often demands specialized low-bit hardware, pruning directly eliminates operations, natively accelerating inference.
Pruning roots trace to second-order methods like OBD~\cite{LeCun1989OBD} and OBS~\cite{Hassibi1992OBS}.
Later approaches introduced magnitude thresholding~\cite{Han2015Mag}, gradient-based sensitivities~\cite{Lee2019SNIP}, and identified sparse, trainable subnetworks~\cite{Frankle2019LTH}.
However, these \emph{unstructured} methods create irregular sparsity requiring custom accelerators~\cite{Gale2019Sparsity}.
\emph{Structured} pruning instead removes entire neurons or channels~\cite{Li2017FilterPrune}, yielding immediate speedups on commodity hardware at the cost of severe accuracy drops, typically requiring extensive iterative retraining~\cite{Yang2023NViT}.

\subsection{Variance-Based Pruning}
Variance-Based Pruning (VBP)~\cite{Berisha2025VBP} selects neurons for removal based on activation variance rather than weight magnitude.
Given an input activation vector $\mathbf{x}$ with mean $\boldsymbol{\mu}$, VBP introduces a mean-shift vector to compensate for pruned neurons. The mean-shift vector $\Delta\boldsymbol{\mu}$'s $j$-th entry is $\boldsymbol{\mu}_j$ if neuron $j$ is pruned and zero otherwise.
The output of the pruned layer is then expressed as:
\begin{equation}
\mathbf{y} = \hat{\mathbf{W}}\hat{\mathbf{x}}+\mathbf{b}^\prime \label{eq:output_update}
\end{equation}
where $\hat{\mathbf{W}}$ is the pruned weight of the subsequent layer, $\hat{\mathbf{x}}$ is the pruned output, and $\mathbf{b}^\prime$ is the mean-shift compensation term that absorbs the effect of the pruned neurons into the bias:
\begin{equation}
\mathbf{b}^\prime = \mathbf{b} + \mathbf{W}\Delta\boldsymbol{\mu} \label{eq:bias_update}.
\end{equation}
After this compensation, the residual reconstruction error for neuron $i$ is governed by its activation variance, $\sigma_i^2$:
\begin{equation}
\mathbb{E}[(x_i - \mathbb{E}[x_i])^2] = \mathbb{E}[(x_i - \mu_i)^2]=\sigma_i^2
\end{equation}
Therefore, pruning neurons with the lowest $\sigma_i^2$ minimizes the expected reconstruction error under bias compensation.

\subsection{Optimal Brain Compression}
The Optimal Brain Compression (OBC) framework~\cite{Frantar2022OBC} has served as the foundation for numerous model compression algorithms, notably extending to Large Language Models through methods such as GPTQ~\cite{Frantar2023GPTQ} and SparseGPT~\cite{Frantar2023SparseGPT}.
Building upon the Optimal Brain Surgeon (OBS)~\cite{Hassibi1992OBS} framework, OBC minimizes the squared reconstruction error between the original and compressed layer outputs:
\begin{equation}
\operatorname*{argmin}_{\mathbf{\hat{W}}}||\mathbf{WX}-\mathbf{\hat{W}X}||^2_F,
\end{equation}
where $\hat{\mathbf{W}}$ is the compressed weight matrix and $\mathbf{X} \in \mathbb{R}^{d \times N}$ is the input activation matrix whose columns are activation vectors collected from $N$ calibration samples.
To minimize this objective, the OBC framework derives a block-wise closed-form update rule that adjusts the remaining weights as
\begin{equation}
\mathbf{\delta}_P = -\mathbf{H}^{-1}_{:,P}((\mathbf{H}^{-1})_P)^{-1}\mathbf{w}_P
\end{equation}
where $\mathbf{\delta}_P$ is the weight update for block $P$, $\mathbf{H} = 2\mathbf{X}\mathbf{X}^T$ is the layer-wise proxy-Hessian, $\mathbf{w}_P$ is the current weight vector of the row being updated, and $P$ denotes the set of indices corresponding to the block being compressed.
Recent work OBS-Diff~\cite{Zhu2026OBSDiff} adapts second-order statistics for structured pruning of diffusion models.
However, OBS-Diff predominantly utilizes complex, Hessian-based OBS calculations as the primary criterion for selecting which structures to remove.
Our method decouples this pipeline: we employ an efficient denoised variance metric to identify redundant neurons, reserving the OBC framework strictly for optimal weight recovery after pruning.

%% file: sections/03_method.tex
\section{Methodology}
\label{sec:method}
In this section, we present the DVBP + $\text{OB}^2\text{C}$ structured pruning pipeline.
We begin by augmenting the OBC~\cite{Frantar2022OBC} reconstruction objective with mean-shift compensation (\S\ref{sec:ob2c}), showing that the resulting layer-wise Hessian reduces to the input activation covariance matrix.
We then describe an online method for computing this covariance without prohibitive memory costs (\S\ref{sec:cov_comp}).
To address the inherent noise in finite-sample covariance estimates, we apply spectral denoising via the Marchenko-Pastur distribution (\S\ref{sec:mp_denoise}), yielding robust variance scores for neuron selection (\S\ref{sec:dvar_score}).
Finally, we detail the end-to-end pruning and weight update mechanism (\S\ref{sec:prune_update}).

\subsection{Optimal Brain Bias Compensation ($\text{OB}^2\text{C}$)}
\label{sec:ob2c}
To mitigate the mean-shift error introduced during weight pruning, we introduce Optimal Brain Bias Compensation ($\text{OB}^2\text{C}$), which explicitly incorporates a bias term into the layer-wise reconstruction objective.
Given an input activation matrix $\mathbf{X} \in \mathbb{R}^{d \times N}$, original weights $\mathbf{W}$, pruned weights $\hat{\mathbf{W}}$, and a bias vector $\mathbf{b}$, the augmented objective is formulated as:
\begin{equation}
\arg\min_{\hat{\mathbf{W}}, \mathbf{b}} \left\| \mathbf{W}\mathbf{X} - (\hat{\mathbf{W}}\mathbf{X} + \mathbf{b}\mathbf{1}^T) \right\|_F^2
\end{equation}
where $\mathbf{1}$ is an $N$-dimensional vector of ones.
Setting the partial derivative of the objective with respect to $\mathbf{b}$ to zero yields the optimal bias compensation:
\begin{equation}
\mathbf{b}^* = (\mathbf{W} - \hat{\mathbf{W}})\boldsymbol{\mu}
\end{equation}
where $\boldsymbol{\mu} = \frac{1}{N}\mathbf{X}\mathbf{1}$ denotes the mean vector of the input activations.
Substituting $\mathbf{b}^*$ back into the objective function yields a simplified minimization problem:
\begin{equation}
\arg\min_{\Delta\mathbf{W}} \left\| \Delta\mathbf{W}\widetilde{\mathbf{X}} \right\|_F^2
\end{equation}
where $\Delta\mathbf{W} = \mathbf{W} - \hat{\mathbf{W}}$ represents the weight perturbation, and $\widetilde{\mathbf{X}} =\mathbf{X} - \boldsymbol{\mu}\mathbf{1}^T$ represents the mean-centered activations. For detailed mathematical derivation, see Appendix \ref{appendix:ob2c}.

Evaluating the error induced by this weight perturbation, we find that the layer-wise proxy-Hessian matrix with respect to the rows of $\Delta\mathbf{W}$ is given by:
\begin{equation}
\mathbf{H} = 2\widetilde{\mathbf{X}}\widetilde{\mathbf{X}}^\top = 2N\mathbf{C}
\end{equation}
where $N$ is the calibration set size and $\mathbf{C}$ is the activation covariance matrix.
Consequently, the layer-wise Hessian is directly proportional to the covariance of the centered activations, rather than the uncentered second moment used in standard formulations.
Integrating this into the OBC multi-weight update, we replace the standard Hessian with $2N\mathbf{C}$.
Let $P$ denote the set of indices corresponding to the pruned neurons.
The update formula for the weights is rewritten as:
\begin{equation}
\mathbf{\delta_W} = - \mathbf{W}_{:, P} \left( [\mathbf{C}^{-1}]_{P, P} \right)^{-1} [\mathbf{C}^{-1}]_{P, :}
\end{equation}
This yields an update expression that is structurally identical to the standard OBC rule, but inherently minimizes the mean-shift error by operating on the centered covariance matrix $\mathbf{C}$.

\subsection{Covariance Matrix Computation}
\label{sec:cov_comp}
To apply $\text{OB}^2\text{C}$, we first require the exact centered covariance matrix of the activations for each layer.
Storing full-precision activations for a naive computation demands an $\mathcal{O}(N)$ memory footprint, rendering the approach computationally infeasible on large calibration sets.
To ensure our approach remains computationally tractable, we track each layer's covariance matrix in an online, batched manner during calibration.
To track the activation statistics of neurons, the VBP~\cite{Berisha2025VBP} paper employs Welford's algorithm to iteratively update their mean and variance.
In a similar vein, Chan et al.~\cite{Chan1982SecondMoment} introduced a robust formula for updating the mean and second central moment across partitioned dataset $\mathbf{A}$ partitioned into $\mathbf{B}$ and $\mathbf{C}$ where $\mathbf{A} = \mathbf{B} \cup \mathbf{C}$ , using the expressions from Pébay~\cite{Pebay2008SecondMoment}:
\begin{equation}
\boldsymbol{\mu_A} = \boldsymbol{\mu}_{\mathbf{B}} + n_{\mathbf{C}}\frac{\delta_{\mathbf{C},\mathbf{B}}}{n}, \ \ \ \ \ \
\mathbf{M}_{2, \mathbf{A}} = \mathbf{M}_{2, \mathbf{B}} + \mathbf{M}_{2, \mathbf{C}} + n_{\mathbf{B}} n_{\mathbf{C}} \frac{\delta_{\mathbf{C},\mathbf{B}}^2}{n_A}
\label{eq:mean}
\end{equation}
where $\mathbf{M}_2$ denotes the second central moment matrix,$n_{\mathbf{A}}$, $n_{\mathbf{B}}$, $n_{\mathbf{C}}$ are the respective cardinalities, and $\delta_{\mathbf{C},\mathbf{B}} = \boldsymbol{\mu}_{\mathbf{C}} - \boldsymbol{\mu_{\mathbf{B}}}$ is the difference between the partition means.

Let $\mathcal{P}$ denote the set of previously processed activations containing $n_\mathcal{P}$ samples, and let $\mathcal{N}$ denote an incoming new batch containing $n_\mathcal{N}$ samples.
The combined set is given by $\mathcal{C} = \mathcal{P} \cup \mathcal{N}$, with a total of $n_\mathcal{C} = n_\mathcal{P} + n_\mathcal{N}$ samples.
When a new batch arrives, we first update the running mean of the activations using Eq.~\ref{eq:mean}. 

To update the sample covariance matrix, we define the weighting factor $\alpha = \frac{n_\mathcal{P}}{n_C}$, which implies $1-\alpha = \frac{n_\mathcal{N}}{n_C}$.
By decomposing the second moment across the disjoint sets $\mathcal{P}$ and $\mathcal{N}$, we derive the following online update formula for the covariance matrix:
\begin{equation}
\mathbf{C}_{\mathcal{C}} = \alpha\mathbf{C}_{\mathcal{P}} + (1-\alpha)\mathbf{C}_{\mathcal{N}} + \alpha(1-\alpha)\boldsymbol{\delta}\boldsymbol{\delta}^T
\end{equation}
where $\boldsymbol{\delta} = \boldsymbol{\mu}_{\mathcal{N}} - \boldsymbol{\mu}_{\mathcal{P}}$ represents the difference between the mean vectors of the new batch and the previously accumulated data.
Using this recursive formulation, we can accurately construct the sample covariance matrix of each layer for an arbitrarily large calibration set without requiring the full activation history to be held in memory.

\subsection{Denoising the Covariance Matrix with Marchenko-Pastur}
\label{sec:mp_denoise}
\subsubsection{Marchenko-Pastur Distribution}
We observe that the spectral density of the activation covariance matrix $\mathbf{C}$ exhibits a bulk profile characteristic of the MP~\cite{Marchenko1967MP} distribution, strongly suggesting that the underlying signal subspace is corrupted by noise.
To mitigate this, we denoise $\mathbf{C}$ by conducting eigendecomposition on $\mathbf{C}$ and by fitting its eigenvalue spectrum to the MP distribution.
The MP distribution describes the asymptotic eigenvalue behavior of large random matrices.
Specifically, for a large random matrix $\mathbf{X}$ of size $m\times n$ whose entries are independent and identically distributed (i.i.d.) with mean 0 and a variance of $\mathbf{\sigma}^2$, the probability density function of the eigenvalues of its corresponding sample covariance $\mathbf{Y}_n = \frac{1}{n}\mathbf{X}\mathbf{X}^T$ is expressed as:
\begin{equation}
f(x) = \frac{\sqrt{(\lambda_{+} - x)(x-\lambda_{-})}}{2\pi\sigma^2\lambda x}
\end{equation}
for $\lambda_{-}<x<\lambda_+$ and $0$ otherwise.
Here, $\lambda= \frac{m}{n}$ represents the aspect ratio of the matrix dimensions.
According to the MP Law, if $\mathbf{X}$ satisfies the conditions, all of its non-zero eigenvalues will asymptotically fall within the theoretical bounds of $\lambda_+$ and $\lambda_-$.
These bounds can be obtained by the following equation
\begin{equation}
\label{eq:lambda+-}
\lambda_\pm = \sigma^2(1 \pm \sqrt{\lambda})^2.
\end{equation}

\subsubsection{Noise Variance of Noisy Matrices}
Gavish and Donoho~\cite{Gavish2014SV} focuses on the analysis of noisy matrices under the additive model $\mathbf{Y} = \mathbf{X}+\sigma\mathbf{Z}$, where $\mathbf{Y}$ is the observed data matrix, $\mathbf{X}$ is the underlying low-rank signal, and $\mathbf{Z}$ represents random noise.
A primary challenge in practical application is that the true noise scale, $\sigma$, is typically unknown.
To address this, the authors formulate a robust estimator $\hat{\sigma}(\mathbf{Y})$ using the ratio of the median singular value of the data matrix to the theoretical median of the MP distribution:
\begin{equation}
\label{eq:sigma_est}
\hat{\sigma}(\mathbf{Y})\stackrel{\text{def}}{=} \frac{y_{med}}{\sqrt{n\mu_{\lambda}}},
\end{equation}
where $y_{\text{med}}$ is the median singular value of $\mathbf{Y}$. $\mu_\lambda$ is the median of MP distribution obtainable by solving for $\lambda_-<x<\lambda_+$
\begin{equation}
\label{eq:integral_med}
\int_{\lambda_-}^{x}\frac{\sqrt{(\lambda_+ - t)(t-\lambda_-)}}{2\pi t} dt = \frac{1}{2}.
\end{equation}
Following our observation where eigenvalue distribution contains MP-like noise, visualized in Fig.~\ref{fig:figure2}.
We assume the centered activation data matrix $\widetilde{\mathbf{X}}$ follows an additive noise model, $\widetilde{\mathbf{X}} = \widetilde{\mathbf{X}}_{\text{signal}} + \sigma\mathbf{Z}$, where the entries of the noise matrix $\mathbf{Z}$ adhere to the Marchenko-Pastur conditions. 

Following Gavish and Donoho's work~\cite{Gavish2014SV}, we estimate the noise scale $\sigma$ by comparing the median eigenvalue of $\mathbf{C}$ to the theoretical median of the MP distribution.
The median singular value of the data matrix, denoted as $s_{\text{med}}$, relates to the median eigenvalue of the covariance matrix, $\Lambda_{\text{med}}$, obtained through eigendecomposition as follows:
\begin{equation}
s_{\text{med}} = \sqrt{N\Lambda_{\text{med}}}.
\end{equation}
Substituting this relationship into the estimator Eq.~\ref{eq:sigma_est} yields:
\begin{equation}
\hat{\sigma}(\widetilde{\mathbf{X}}) = \sqrt{\frac{\Lambda_{\text{med}}}{\mu_{\lambda}}}
\end{equation}
where $\mu_{\lambda}$ is the theoretical median of the MP distribution. 

To obtain $\mu_{\lambda}$, we require matrix $\mathbf{\widetilde{X}}$'s aspect ratio $\lambda$.
In this context, we define the aspect ratio as $\lambda = \frac{d}{N}$, where $d$ is the number of input neurons and $N$ is the calibration set size.
We use calibration set size (number of independent images) rather than the total number of patches for the denominator because patches within a single image are highly correlated, using the total patch count would violate the i.i.d. noise assumption.
By defining our sample size strictly as the calibration set size, we ensure the i.i.d. assumption holds.
With the aspect ratio $\lambda$, we compute the theoretical median eigenvalue $\mu_{\lambda}$ via Eq.~\ref{eq:integral_med}, which subsequently allows us to estimate the noise scale $\sigma$.

Using the estimated noise variance, we calculate the theoretical MP upper bound $\lambda_+$ using Eq.~\ref{eq:lambda+-}.
After computing the eigendecomposition of the covariance matrix $\mathbf{C} = \mathbf{V}\mathbf{\Lambda}\mathbf{V}^T$, where $\mathbf{\Lambda} = \operatorname{diag}(\lambda_1, \dots, \lambda_d)$ contains the eigenvalues and $\mathbf{V} = [\mathbf{v}_1, \dots, \mathbf{v}_d]$ the corresponding eigenvectors, we apply hard thresholding to retain only the signal components:
\begin{align}
\mathbf{\Lambda}_{\text{signal}} &= \operatorname{diag}(\lambda_i) \quad \forall \lambda_i > \lambda_+ \\
\mathbf{V}_{\text{signal}} &= [\mathbf{v}_i] \quad \forall \lambda_i > \lambda_+
\end{align}

\subsection{Denoised Variance Score}
\label{sec:dvar_score}
Using the retained signal components, we reconstruct the denoised covariance matrix:
\begin{equation}
\mathbf{C}_{\text{denoised}} = \mathbf{V}_{\text{signal}}\mathbf{\Lambda}_{\text{signal}}\mathbf{V}_{\text{signal}}^T
\end{equation}
Finally, we extract the denoised variance scores for each neuron directly from the diagonal entries of the reconstructed matrix:
\begin{equation}
\mathbf{s}_{\text{dvar}} = \operatorname{diag}(\mathbf{C}_{\text{denoised}})
\end{equation}
Empirically, we observe that the denoised variance scores $\mathbf{s}_{\text{dvar}}$ excel at high pruning ratios, whereas the raw variance scores $\mathbf{s}_{\text{var}} = \operatorname{diag}(\mathbf{C})$, extracted from the original covariance matrix, remain highly effective at low pruning ratios.
Thus, we store the raw scores prior to eigendecomposition and formulate a hybrid scoring metric, controlled by the target pruning ratio $p$:
\begin{equation}
\mathbf{s}_{\text{hybrid}} = (1-p)\mathbf{s}_{\text{var}} + p\mathbf{s}_{\text{dvar}}
\end{equation}

\subsection{Pruning and Update Mechanism}
\label{sec:prune_update}
Following the structural framework of Variance-Based Pruning (VBP), we target MLP blocks in Vision Transformers (ViTs)~\cite{Dosovitskiy2021ViT} and ConvNeXt~\cite{Liu2022ConvNeXt} models.
While we adopt this standard layer selection protocol, our approach introduces a novel neuron selection criterion and weight update mechanism. 

Specifically, we globally rank the hidden dimensions based on our proposed hybrid score, and select the lowest-ranking $p$\% of neurons across the entire network to consist the pruned indices set $P$. Rather than enforcing a uniform per-layer pruning ratio, the network dynamically allocates capacity, preserving critical layers while heavily penalizing redundant ones. 

To mitigate immediate performance degradation from neuron removal, we apply a two-stage compensation. First, we apply our novel $\text{OB}^2\text{C}$ update to the outgoing weight matrix of the block (the second linear layer) by adding $\mathbf{\delta_W}$ computed by by the pruned indices set $P$ to the weights $\mathbf{W}$. Then, we apply mean-shift compensation to the subsequent layer, which is the optimal bias term $\mathbf{b}^*$ added to the existing bias term.

After the two-stage compensation, we apply pruning. For a standard two-layer block, pruning reduces the intermediate hidden dimension from $d_{\text{hidden}}$ to $d_{\text{pruned}}$.
Consequently, the weight matrix of the first layer is reduced from $d_{\text{in}} \times d_{\text{hidden}}$ to $d_{\text{in}} \times d_{\text{pruned}}$, and the weight matrix of the second layer is reduced from $d_{\text{hidden}} \times d_{\text{out}}$ to $d_{\text{pruned}} \times d_{\text{out}}$. Because the external input ($d_{\text{in}}$) and output ($d_{\text{out}}$) dimensions remain strictly preserved, the pruned block seamlessly integrates back into the overarching network architecture. 

%% file: sections/04_experiments_test_results.tex
\section{Experiments}
\label{sec:exp}
We evaluate our DVBP+$\text{OB}^2\text{C}$ structured pruning framework on the Tiny, Small, and Base variants of the DeiT~\cite{Touvron2021DeiT}, Swin~\cite{Liu2021Swin}, and ConvNeXt~\cite{Liu2022ConvNeXt} architectures, all pre-trained on the ImageNet-1K~\cite{deng2009imagenet} dataset.
To compute the activation covariance matrices, we construct a calibration dataset of 8,192 images, randomly sampled from the ImageNet-1K training split using a fixed seed.
Notably, we omit data augmentation during calibration to accurately capture the true, unperturbed neuron activation statistics.
Our evaluation primarily focuses on direct post-pruning accuracy retention, aiming to maximize off-the-shelf performance without relying on expensive retraining or fine-tuning pipelines.
All experiments are conducted on a single NVIDIA GeForce RTX 3090 GPU.
To guarantee a rigorous and fair comparison, we locally compute both the uncompressed baseline accuracies and the post-pruning performance of all evaluated methods under the exact same setup.
\begin{table}[t]
\centering
\caption{Comparison of Top-1 accuracy ($\%$) on ImageNet-1K across different pruning methods on DeiT and Swin.
All models are evaluated immediately after compression without any post-pruning fine-tuning.}
\label{tab:table1}
\includegraphics[width=\textwidth, keepaspectratio]{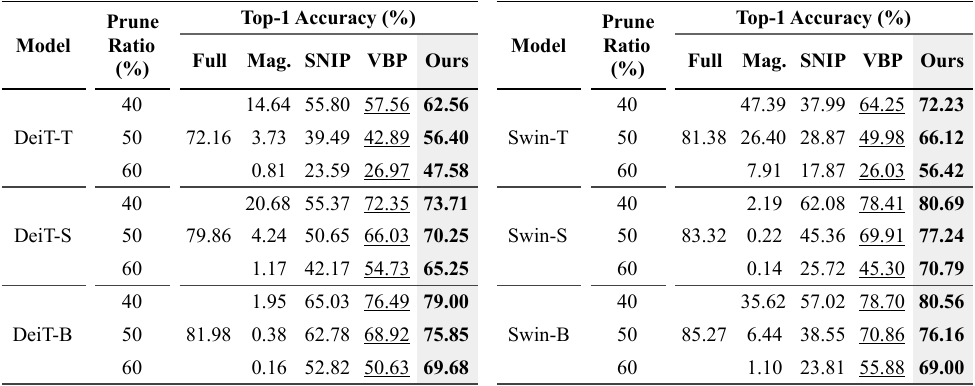}
\end{table}

\begin{table}[t]
\centering
\caption{Comparison of parameter counts (M), MACs (G), and Top-1 accuracy ($\%$) for DeiT models at a 50$\%$ pruning ratio.}
\label{tab:table2}
\includegraphics[width=\textwidth, keepaspectratio]{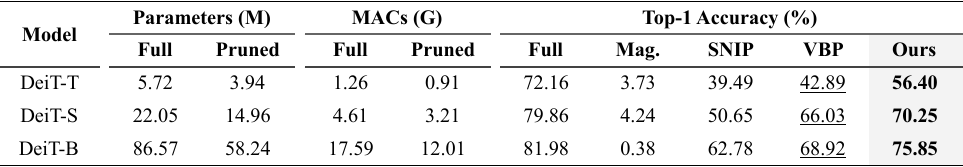}
\end{table}

\begin{figure}[t]
\centering
% --- Left Figure ---
\begin{minipage}{0.48\textwidth}
  \centering
  \includegraphics[width=\linewidth]{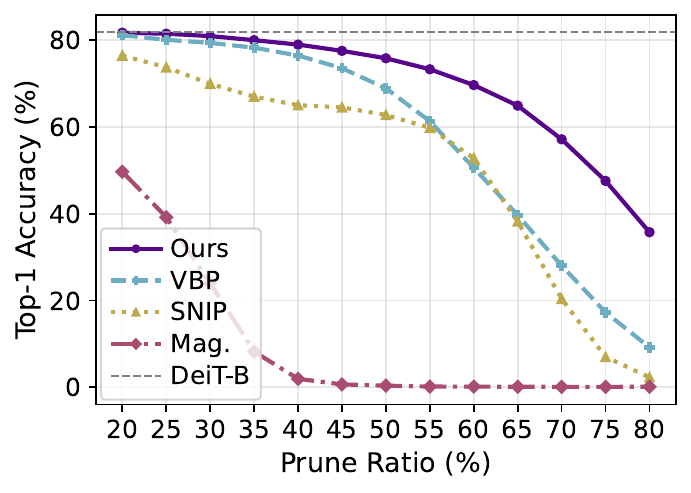}
\end{minipage}\hfill
% --- Right Figure ---
\begin{minipage}{0.48\textwidth}
  \centering
  \includegraphics[width=\linewidth]{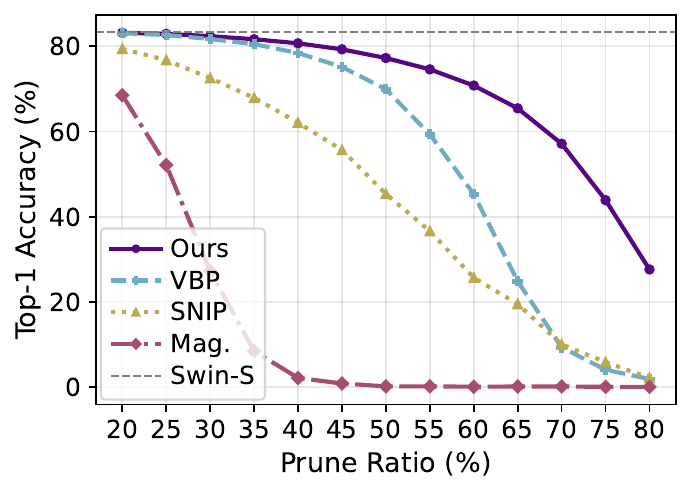}
\end{minipage}

\caption{Pruning accuracy across diverse pruning ratios for DeiT-B (left) and Swin-S (right).
SNIP and Magnitude pruning are adjusted to structured pruning.}
\label{fig:figure1}
\end{figure}

\subsection{Results}
Our results demonstrate that the proposed approach consistently outperforms all baseline pruning methods across all evaluated models and pruning ratios.
Notably, when compared to our closest competitor, VBP~\cite{Berisha2025VBP}, the accuracy gap widens as the pruning ratio increases.
For instance, on the Swin-S model, the post-pruning accuracy gap between our method and VBP is 2.28$\%$ at a 40$\%$ pruning ratio.
This margin increases substantially to 7.33$\%$ at a 50$\%$ ratio, and expands even further to 25.49$\%$ at a 60$\%$ ratio.

As detailed in Table~\ref{tab:table1}, at a 50$\%$ pruning ratio, our method successfully retains approximately 80$\%$ of the original accuracy for the Tiny models, and roughly 90$\%$ for the Small and Base models, representing a significant improvement over VBP.
This diverging performance trend is further illustrated in Fig.~\ref{fig:figure1}, which shows the performance gap between our approach and VBP becoming increasingly pronounced from the 40$\%$ pruning threshold onward.

Finally, because our approach targets the same layers and utilizes the same fundamental pruning mechanism as the baselines, the overall parameter counts and multiply-accumulate operations (MACs) remain identical.
However, despite this equivalent computational footprint, our method yields a vastly superior top-1 accuracy (Table~\ref{tab:table2}).

\begin{figure}[t]
\centering
% --- Left Figure ---
\begin{minipage}[b]{0.23\textwidth}
  \centering
  \includegraphics[width=\linewidth]{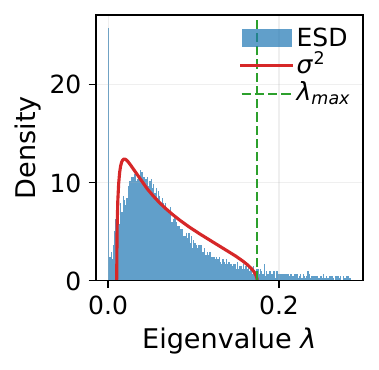}
  \centerline{(a) DeiT-B Layer 9} % Sub-caption
\end{minipage}\hfill
%left middle
\begin{minipage}[b]{0.23\textwidth}
  \centering
  \includegraphics[width=\linewidth]{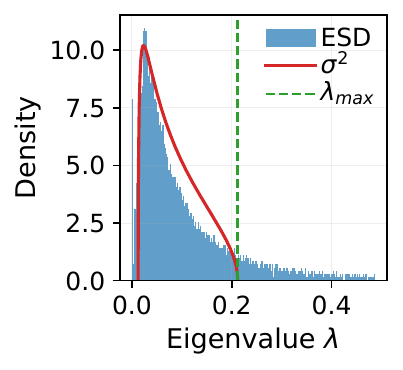}
  \centerline{(b) DeiT-B Layer 10} % Sub-caption
\end{minipage}\hfill
%left middle
\begin{minipage}[b]{0.23\textwidth}
  \centering
  \includegraphics[width=\linewidth]{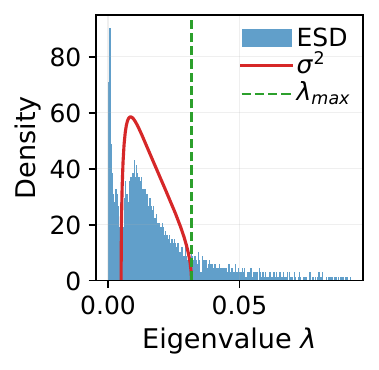}
  \centerline{(c) Swin-S Layer 11} % Sub-caption
\end{minipage}\hfill
% --- Right Figure ---
\begin{minipage}[b]{0.23\textwidth}
  \centering
  \includegraphics[width=\linewidth]{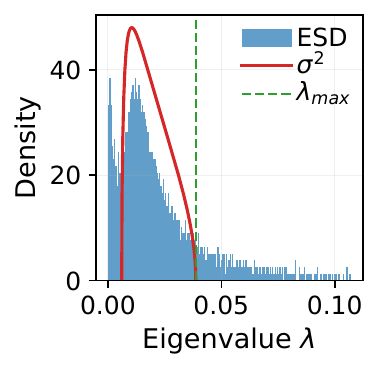}
  \centerline{(d) Swin-S Layer 12} % Sub-caption
\end{minipage}

\caption{Empirical spectral density fitted with the Marchenko-Pastur distribution.
The largest 5\% of eigenvalues are excluded for visual clarity.}
\label{fig:figure2}
\end{figure}

As illustrated in Fig.~\ref{fig:figure2}, the empirical eigenvalue distributions of the models exhibit a noise structure characteristic of the Marchenko-Pastur (MP)~\cite{Marchenko1967MP} distribution.
Furthermore, the fitted MP~\cite{Marchenko1967MP} curve accurately captures these noise components.

\begin{figure}[t]
\centering
% --- Left Figure ---
\begin{minipage}{0.48\textwidth}
  \centering
  \includegraphics[width=\linewidth]{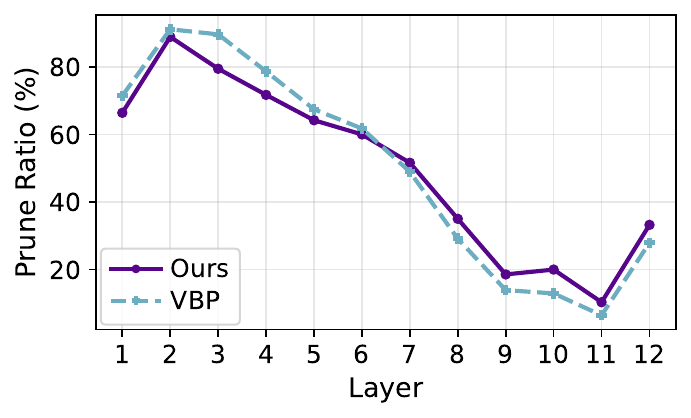}
\end{minipage}\hfill
% --- Right Figure ---
\begin{minipage}{0.48\textwidth}
  \centering
  \includegraphics[width=\linewidth]{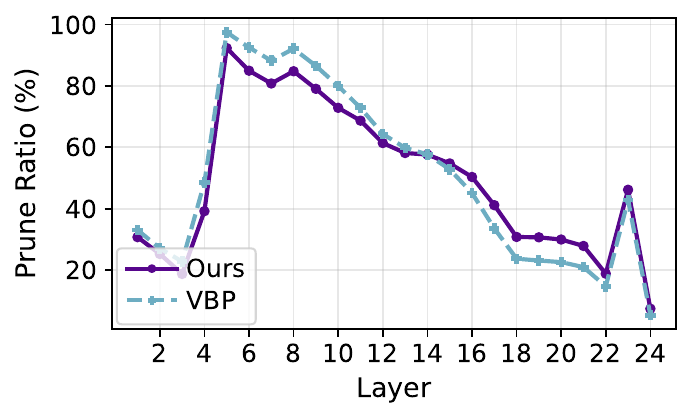}
\end{minipage}

\caption{Layerwise pruning ratio for DeiT-B (left) and Swin-S (right), for model pruning rate 50$\%$.}
\label{fig:figure3}
\end{figure}

\begin{figure}[t]
\centering
% --- Left Figure ---
\begin{minipage}{0.48\textwidth}
  \centering
  \includegraphics[width=\linewidth]{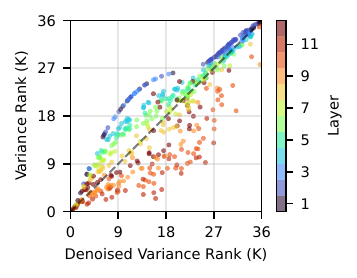}
\end{minipage}\hfill
% --- Right Figure ---
\begin{minipage}{0.48\textwidth}
  \centering
  \includegraphics[width=\linewidth]{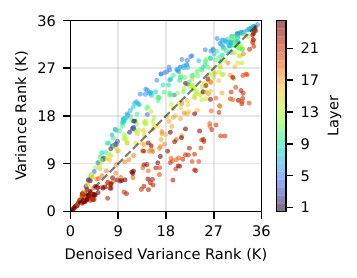}
\end{minipage}

\caption{Spearman correlation plot of DeiT-B (left) and Swin-S (right) between denoised and raw variance score rank.}
\label{fig:figure4}
\end{figure}

As shown in Fig.~\ref{fig:figure3}, our hybrid scoring mechanism yields a distinct pruning profile compared to VBP~\cite{Berisha2025VBP}.
Specifically, our approach prunes the earlier layers less aggressively while increasing the pruning rate in the deeper layers.
However, because both methods rely on a variance-based metric, their overall macroscopic trends remain similar.
This relationship is further quantified by the Spearman rank correlation plot (Fig.~\ref{fig:figure4}), which reveals a general linear correlation between the ranks of the standard variance and our denoised scores, albeit with considerable deviations occurring within the middle portion of the rankings.

\begin{figure}[t]
  \centering
  % --- Left Column: The Table ---
  \begin{minipage}{0.48\textwidth}
    \centering
    \captionof{table}{Results of ConvNeXt Pruning at 50$\%$.}
    \label{tab:table3}
    \includegraphics[width=\linewidth, keepaspectratio]{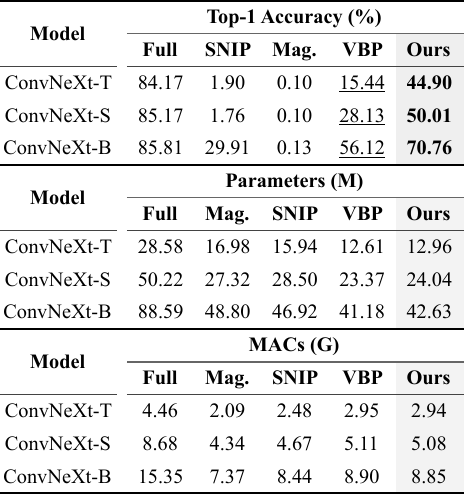}
  \end{minipage}
  \hfill % This creates the space between the two columns
  % --- Right Column: The Figure ---
  \begin{minipage}{0.48\textwidth}
    \centering
    \includegraphics[width=\linewidth]{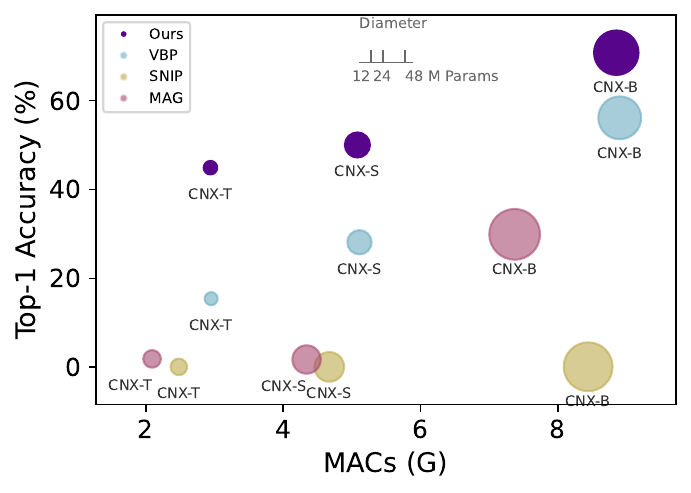}
    \caption{Overall Diagram for ConvNeXt pruning results.}
    \label{fig:figure5}
  \end{minipage}
\end{figure}

As demonstrated in Table~\ref{tab:table3} and Fig.~\ref{fig:figure5}, our pruning methodology generalizes effectively to the ConvNeXt~\cite{Liu2022ConvNeXt} Tiny, Small, and Base models.
SNIP~\cite{Lee2019SNIP} and Magnitude~\cite{Han2015Mag} pruning cause severe performance degradation on ConvNeXt architectures, reducing the accuracy to chance-level performance on the ImageNet-1K dataset.
While VBP does achieve some accuracy retention, it is considerably lower than the baseline accuracies.
Our method improves upon VBP by 29.46$\%$ for ConvNeXt-T, 21.88$\%$ for ConvNeXt-S, and 14.64$\%$ for ConvNeXt-B, representing a substantial improvement.
This demonstrates that our method is highly applicable to diverse model architectures.

\subsection{Analysis and Ablation Studies}
\begin{table}[t]
\centering
% --- Left Table (Table 1) ---
\begin{minipage}[b]{0.48\textwidth} % Added [b] here
  \centering
  \caption{Memory and VRAM usage comparison between VBP.}
  \label{tab:table4}
  \includegraphics[width=\linewidth, keepaspectratio]{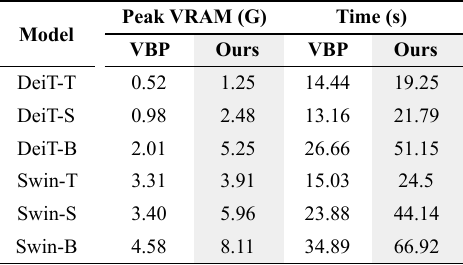}
\end{minipage}\hfill
% --- Right Table (Table 2) ---
\begin{minipage}[b]{0.48\textwidth} % Added [b] here
  \centering
  \caption{Ablation study: denoising and $\text{OB}^2\text{C}$.}
  \label{tab:table5}
  \includegraphics[width=\linewidth, keepaspectratio]{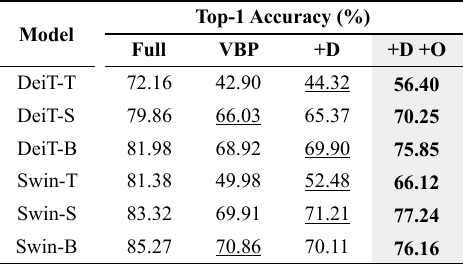}
\end{minipage}
\end{table}

\begin{figure}[t]
\centering
% --- Left Figure ---
\begin{minipage}{0.48\textwidth}
  \centering
  \includegraphics[width=\linewidth]{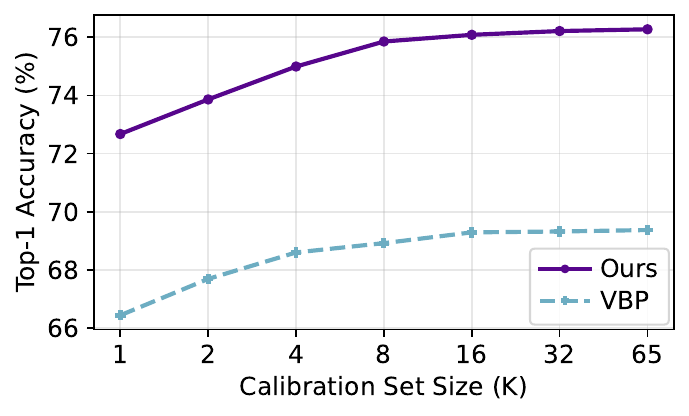}
\end{minipage}\hfill
% --- Right Figure ---
\begin{minipage}{0.48\textwidth}
  \centering
  \includegraphics[width=\linewidth]{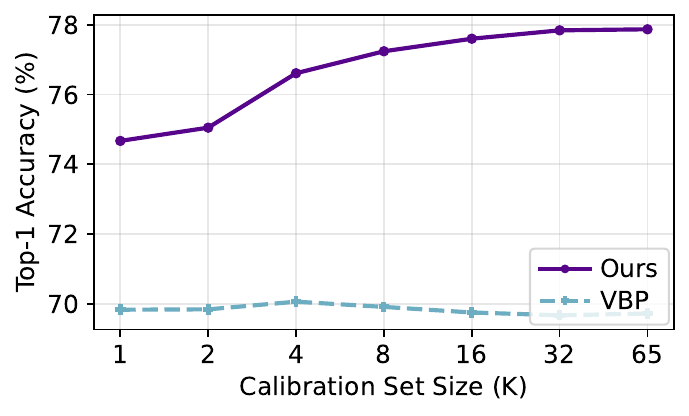}
\end{minipage}

\caption{Top1-Accuracy ($\%$) across various calibration set sizes for DeiT-B (left) and Swin-S (right).}
\label{fig:figure6}
\end{figure}

Despite the significant accuracy gains, our method introduces minimal computational and memory overhead, ensuring its practical deployment on consumer-grade GPUs.
Because we utilize a block-wise approach in $\text{OB}^2\text{C}$ and precompute the pruning selections via our hybrid scoring mechanism, peak VRAM consumption is strictly bounded under 10 GB.
Furthermore, the pruning process is highly time-efficient; as detailed in Table~\ref{tab:table4}, while the execution time is approximately $2\times$ that of VBP~\cite{Berisha2025VBP}, it completes in under one minute for most architectures, requiring only 66.92 seconds for the larger Swin-B~\cite{Liu2021Swin} model.
These constrained resource requirements demonstrate that our approach can readily scale to larger models without necessitating specialized, high-memory hardware.

To evaluate the individual contributions of our proposed components, we analyze the isolated performance of the denoising mechanism and the $\text{OB}^2\text{C}$ compensation framework, as detailed in Table~\ref{tab:table5}.
Integrating the denoised scoring yields marginal but consistent accuracy gains of 1-2$\%$ for the majority of the evaluated models, with the exception of DeiT-S~\cite{Touvron2021DeiT} and Swin-B~\cite{Liu2021Swin}, which experience a negligible decrease of less than 1$\%$.
In contrast, the application of the $\text{OB}^2\text{C}$ framework serves as the primary driver of performance recovery.
Notably, it delivers a substantial absolute accuracy improvement of nearly 14$\%$ for DeiT-T~\cite{Touvron2021DeiT}, while consistently boosting the performance of the remaining models by 4-7$\%$.

Because both VBP~\cite{Berisha2025VBP} and our proposed method can accommodate an arbitrary calibration set size, we investigate how varying this size impacts the final top-1 accuracy of each approach.
While our primary experiments utilize a calibration dataset of 8,192 samples, our proposed method demonstrates robust scalability with respect to calibration size.
As illustrated in Figure~\ref{fig:figure6}, increasing the calibration set initially yields top-1 accuracy improvements for DeiT-B pruning for both our method and VBP, with both approaches reaching convergence at approximately 8K samples.
However, for Swin-S~\cite{Liu2021Swin}, our method exhibits continuous performance gains as the calibration set expands, achieving convergence at roughly 32K samples.
In stark contrast, the performance of VBP plateaus entirely in this scenario, showing no measurable improvement across a sweep from 1K to 65K samples.

\begin{table}[t]
\centering
\caption{Ablation Study: comparing various denoising methods.}
\label{tab:table6}
\includegraphics[width=1\textwidth, keepaspectratio]{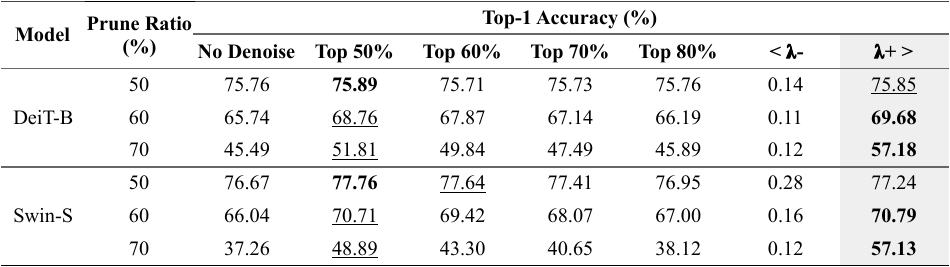}
\end{table}

\begin{figure}[t]
\centering
% --- Left Figure ---
\begin{minipage}{0.48\textwidth}
  \centering
  \includegraphics[width=\linewidth]{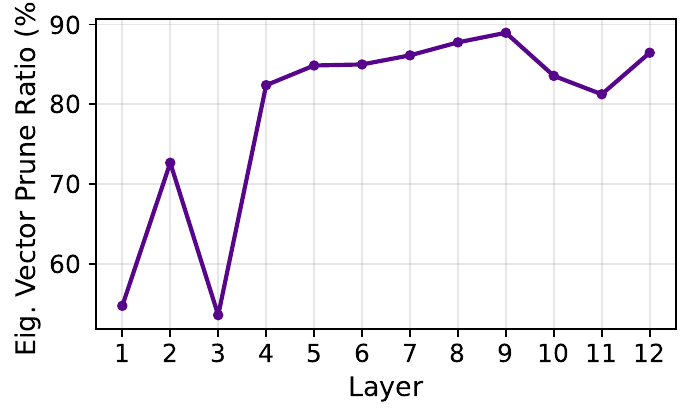}
  \centerline{(a) DeiT-B} % Sub-caption
\end{minipage}\hfill
% --- Right Figure ---
\begin{minipage}{0.48\textwidth}
  \centering
  \includegraphics[width=\linewidth]{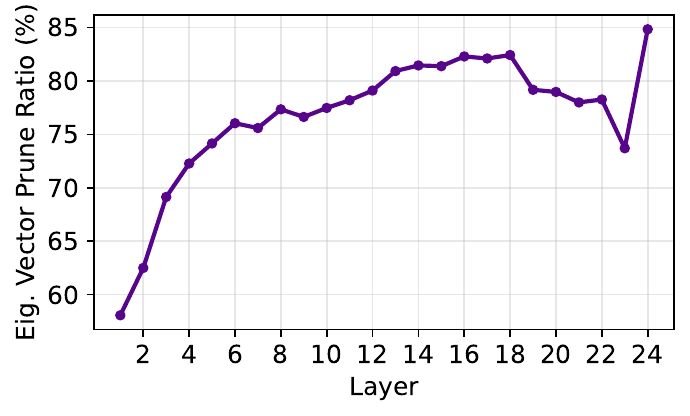}
  \centerline{(b) Swin-S} % Sub-caption
\end{minipage}

\caption{Eigenvector prune ratio based on Marchenko-Pastur denoising.}
\label{fig:figure7}
\end{figure}

We systematically compare various denoising strategies to validate our approach of fitting the MP~\cite{Marchenko1967MP} distribution and retaining only the signal eigenvectors associated with eigenvalues larger than $\lambda_+$.
As demonstrated in Table~\ref{tab:table6}, isolating eigenvectors with eigenvalues exceeding $\lambda_+$ consistently yields the highest top-1 accuracy among the evaluated denoising methods.
Furthermore, this analysis confirms that eigenvectors with eigenvalues below $\lambda_-$ predominantly capture noise, as reconstructing representations solely from these components degrades performance to chance-level accuracy. To quantify the extent of eigenvector pruning performed by our MP fitting method, we visualize the layer-wise pruning ratios in Figure~\ref{fig:figure7}.
The results indicate that our method dynamically adjusts the pruning severity according to network depth; it discards approximately 60-70$\%$ of eigenvectors in the early layers and increases this ratio to roughly 80-90$\%$ in the deeper layers, a trend consistently observed in both the DeiT-B~\cite{Touvron2021DeiT} and Swin-S~\cite{Liu2021Swin} models.
This adaptive behavior supports the hypothesis that representations become increasingly low-rank as they propagate deeper into the network, thereby requiring fewer signal eigenvectors to accurately capture the underlying information.

%% file: sections/05_conclusion.tex
\section{Conclusion}
\label{sec:con}
In this paper, we introduce DVBP + $\text{OB}^2\text{C}$, a novel structured pruning methodology that integrates statistical denoising techniques—specifically, Marchenko-Pastur (MP) distribution fitting within the VBP framework—with a highly efficient compensation mechanism.
Our $\text{OB}^2\text{C}$ framework advances the standard Optimal Brain Compression (OBC) approach by incorporating a multi-weight update strategy for pruned weights, further refined by mean-shift compensation.
Extensive evaluations demonstrate that our method achieves substantial improvements in post-pruning accuracy without the need for subsequent fine-tuning or retraining.
Because this approach is fundamentally applicable to any linear layer, it generalizes seamlessly across a wide range of modern deep neural network architectures.
We hope this work encourages further exploration into structured pruning within the computer vision community, paving the way for near-lossless compression combined with significant inference acceleration.

%% file: sections/A_appendix.tex
\clearpage
\appendix
%% After \appendix, section numbers switch to letters (A, B, ...).
%% Put proofs, extra results, derivations, or implementation notes here.
%% \clearpage above makes the appendix always start on a fresh page.

\section{Additional Details}
\label{sec:intro}

In this supplementary material, the following additional details are provided:
\begin{itemize}
    \item DVBP + $\text{OB}^2\text{C}$ Algorithm
    \item Mathematical Proof of $\text{OB}^2\text{C}$
    \item Calibration Set Experiments
    \item Additional Experiments
\end{itemize}

\subsection{DVBP + $\text{OB}^2\text{C}$ Algorithm}

In this section, we provide the step-by-step pseudocode for our combined DVBP and $\text{OB}^2\text{C}$ methodology.

\begin{breakablealgorithm}
\caption{DVBP Pruning with $\text{OB}^2\text{C}$ Compensation}
\label{alg:dvbp_ob2c}
\begin{algorithmic}[1]

\Require Pretrained model, calibration dataset $\mathcal{D}$, pruning ratio $p$

\State Compute layer-wise centered covariance matrices using Algorithm \ref{alg:dvbp_covariance}
\State Denoise covariance matrices and compute hybrid scores using Algorithm \ref{alg:denoise_cov}
\State Prune neurons and update weights using Algorithm \ref{alg:dvbp_ob2c_prune}

\State \Return Pruned model

\end{algorithmic}
\end{breakablealgorithm}

\vspace{8pt}

\begin{breakablealgorithm}
\caption{Online Covariance Update per MLP Layer}
\label{alg:dvbp_covariance}
\begin{algorithmic}[1]

\Require Calibration dataset $\mathcal{D}$, pretrained model

\State Initialize covariance matrix $\mathbf{C} \gets 0$, mean vector $\boldsymbol{\mu} \gets 0$, total samples $n \gets 0$

\For{$i = 1 \dots \text{total batches}$}

    \State Obtain batch activations $\mathbf{x}$
    \State Compute batch mean $\boldsymbol{\mu}_{\mathcal{N}}$ and batch size $n_{\mathcal{N}}$
    \State $\widetilde{\mathbf{x}} \gets \mathbf{x} - \boldsymbol{\mu}_{\mathcal{N}}$

    \State $\mathbf{C}_{\mathcal{N}} \gets \frac{\widetilde{\mathbf{x}}^T \widetilde{\mathbf{x}}}{n_{\mathcal{N}}}$

    \If{$n == 0$}

        \State $\mathbf{C} \gets \mathbf{C}_{\mathcal{N}}$
        \State $\boldsymbol{\mu} \gets \boldsymbol{\mu}_{\mathcal{N}}$
        \State $n \gets n_{\mathcal{N}}$

    \Else

        \State $n_{\mathcal{P}} \gets n$
        \State $\boldsymbol{\mu}_{\mathcal{P}} \gets \boldsymbol{\mu}$

        \State $n \gets n_{\mathcal{P}} + n_{\mathcal{N}}$

        \State $\alpha \gets \frac{n_{\mathcal{P}}}{n}$
        \State $\boldsymbol{\delta} \gets \boldsymbol{\mu}_{\mathcal{N}} - \boldsymbol{\mu}_{\mathcal{P}}$

        \State $\mathbf{C} \gets \alpha \mathbf{C} + (1-\alpha)\mathbf{C}_{\mathcal{N}} + \alpha(1-\alpha)\boldsymbol{\delta}\boldsymbol{\delta}^T$

        \State $\boldsymbol{\mu} \gets \boldsymbol{\mu}_{\mathcal{P}} + (1-\alpha)\boldsymbol{\delta}$

    \EndIf

\EndFor

\State \Return $\mathbf{C}, \boldsymbol{\mu}$

\end{algorithmic}
\end{breakablealgorithm}

\vspace{8pt}

\begin{breakablealgorithm}
\caption{Denoise Covariance Matrix and Obtain Hybrid Scores}
\label{alg:denoise_cov}
\begin{algorithmic}[1]

\Require Covariance matrix $\mathbf{C}$, mean vector $\boldsymbol{\mu}$, total samples $n$

\State $\mathbf{s}_{\text{var}} \gets \operatorname{diag}(\mathbf{C})$
\State $[\mathbf{V}, \mathbf{\Lambda}] \gets \operatorname{eig}(\mathbf{C})$
\State $d \gets$ number of neurons
\State $\gamma \gets \frac{d}{n}$
\State $\lambda_{\pm} \gets (1 \pm \sqrt{\lambda})^2$
\State Obtain $\mu_{\lambda}$ by solving
\[
\int_{\lambda_-}^{x}
\frac{\sqrt{(\lambda_+ - t)(t-\lambda_-)}}{2\pi t} dt = \frac{1}{2}
\]
\State $\lambda_{\text{med}} \gets$ median eigenvalue of $\mathbf{\Lambda}$
\State $\hat{\sigma} \gets \sqrt{\frac{\lambda_{\text{med}}}{\mu_{\lambda}}}$
\State $\lambda_+ \gets \hat{\sigma}^2 (1+\sqrt{\lambda})^2$
\State $\mathbf{\Lambda}_{\text{signal}} \gets \operatorname{diag}(\lambda_i)$ for all $\lambda_i > \lambda_+$
\State $\mathbf{V}_{\text{signal}} \gets [\mathbf{v}_i]$ for all $\lambda_i > \lambda_+$
\State $\mathbf{C}_{\text{denoised}} \gets
\mathbf{V}_{\text{signal}}\mathbf{\Lambda}_{\text{signal}}\mathbf{V}_{\text{signal}}^T$
\State $\mathbf{s}_{\text{dvar}} \gets \operatorname{diag}(\mathbf{C}_{\text{denoised}})$
\State $\mathbf{s}_{\text{hybrid}} \gets
(1-p)\mathbf{s}_{\text{var}} + p\mathbf{s}_{\text{dvar}}$
\State $\mathbf{C}_{\text{denoised}} \gets
\mathbf{C}_{\text{denoised}} +
\left(
\delta \frac{\operatorname{Tr}(\mathbf{C}_{\text{denoised}})}{d}
\right) I$
\State $\mathbf{C}^{-1} \gets \mathbf{C}_{\text{denoised}}^{-1}$
\State \Return $\mathbf{C}^{-1}, \mathbf{s}_{\text{hybrid}}$
\end{algorithmic}
\end{breakablealgorithm}

\vspace{8pt}

\begin{breakablealgorithm}
\caption{DVBP Pruning with $\text{OB}^2\text{C}$ Compensation}
\label{alg:dvbp_ob2c_prune}
\begin{algorithmic}[1]

\Require Hybrid scores $\mathbf{s}_{\text{hybrid}}$, inverse covariance $\mathbf{C}^{-1}$, mean $\boldsymbol{\mu}$, pruning ratio $p$

\State Concatenate all block scores to obtain global scores $\mathbf{s}_{\text{global}}$

\State $\pi \gets \operatorname{argsort}(\mathbf{s}_{\text{global}})$

\State $k \gets p \cdot |\mathbf{s}_{\text{global}}|$

\State $\mathcal{P}_{\text{global}} \gets \{\pi_1,\dots,\pi_k\}$

\For{each MLP block $l$}

    \State Map global indices to block indices:
    $\mathcal{P}_l$

    \State $\mathcal{K}_l \gets \{1,\dots,d_l\} \setminus \mathcal{P}_l$

    \State $\mathbf{W1}_{\text{pruned}} \gets \mathbf{W1}_{\mathcal{K}_l,:}$

    \State $\mathbf{b1}_{\text{pruned}} \gets \mathbf{b1}_{\mathcal{K}_l}$

    \State $\mathbf{W2}_{\text{new}}, \mathbf{b2}_{\text{new}}
    \gets$ $\text{OB}^2\text{C}$ update
    $(\mathbf{W2},\mathbf{b2},\mathbf{C}^{-1},\boldsymbol{\mu},\mathcal{P}_l)$

    \State $\mathbf{W2}_{\text{pruned}} \gets [\mathbf{W2}_{\text{new}}]_{:, \mathcal{K}_l}$

\EndFor

\end{algorithmic}
\end{breakablealgorithm}

\vspace{8pt}

\begin{breakablealgorithm}
\caption{$\text{OB}^2\text{C}$ Update}
\label{alg:ob2c_update}
\begin{algorithmic}[1]

\Require Weights $\mathbf{W}$, bias $\mathbf{b}$, inverse covariance $\mathbf{C}^{-1}$, mean $\boldsymbol{\mu}$, pruned indices $\mathcal{P}$

\State $\mathbf{A} \gets [\mathbf{C}^{-1}]_{\mathcal{P},\mathcal{P}}$

\State $\mathbf{B} \gets [\mathbf{C}^{-1}]_{\mathcal{P},:}$

\State $\mathbf{W}_{\mathcal{P}} \gets \mathbf{W}_{:,\mathcal{P}}$

\State $\Delta \mathbf{W}
\gets
-
\mathbf{W}_{\mathcal{P}}
\mathbf{A}^{-1}
\mathbf{B}$

\State $\mathbf{W}_{\text{new}} \gets \mathbf{W} + \Delta\mathbf{W}$

\State $\mathbf{b}_{\text{new}} \gets
\mathbf{b} - \Delta\mathbf{W}\boldsymbol{\mu}$

\State \Return $\mathbf{W}_{\text{new}}, \mathbf{b}_{\text{new}}$

\end{algorithmic}
\end{breakablealgorithm}

\subsection{Mathematical Proof of $\text{OB}^2\text{C}$} \label{appendix:ob2c}

In this section, we introduce the bias term into the layer-wise objective function as follows:
\begin{equation} \label{eq:obj_joint}
\arg\min_{\hat{\mathbf{W}}, \mathbf{b}} \left\| \mathbf{W}\mathbf{X} - (\hat{\mathbf{W}}\mathbf{X} + \mathbf{b}\mathbf{1}^T) \right\|_F^2.
\end{equation}
To solve this joint optimization problem, we first derive the optimal bias $\mathbf{b}^*$ for any given weight perturbation, which allows us to substitute it back into the objective and optimize solely for $\hat{\mathbf{W}}$. 

Let $\mathbf{W} \in \mathbb{R}^{d_{\text{out}}\times d_{\text{in}}}$ be the weight matrix of a single layer and $\mathbf{X} \in \mathbb{R}^{d_{\text{in}} \times N}$ be the input activation data, where $N$ is the calibration set size. The reconstruction error induced by pruning is:
\begin{equation}
E = \left\| \mathbf{W}\mathbf{X} - (\hat{\mathbf{W}}\mathbf{X} + \mathbf{b}\mathbf{1}^T) \right\|_F^2.
\end{equation}
To find the optimal bias term, we take the derivative of $E$ with respect to the vector $\mathbf{b}$ and set it to zero:
\begin{equation}
\frac{\partial E}{\partial \mathbf{b}} = -2 \sum_{i=1}^{N} (\mathbf{W}\mathbf{x}_i - \mathbf{\widehat{W}}\mathbf{x}_i - \mathbf{b}) = \mathbf{0}.
\end{equation}
This can be rewritten as:
\begin{equation}
\sum_{i=1}^{N} (\mathbf{W}\mathbf{x}_i - \mathbf{\widehat{W}}\mathbf{x}_i) - N\mathbf{b} = \mathbf{0}.
\end{equation}
Solving for $\mathbf{b}$ yields the optimal bias vector $\mathbf{b}^*$:
\begin{equation}
\mathbf{b}^* = (\mathbf{W} - \mathbf{\widehat{W}})\frac{1}{N}\sum_{i=1}^{N}\mathbf{x}_i = (\mathbf{W} - \mathbf{\widehat{W}})\boldsymbol{\mu} = \Delta\mathbf{W}\boldsymbol{\mu}.
\end{equation}
Substituting this optimal bias $\mathbf{b}^*$ back into the original error term $E$:
\begin{equation}
    \begin{aligned}
    E &= \left\| \mathbf{W}\mathbf{X} - (\widehat{\mathbf{W}}\mathbf{X} + \mathbf{b}^*\mathbf{1}^T) \right\|_F^2 \\
    &= \left\|\mathbf{WX} - (\widehat{\mathbf{W}}\mathbf{X} + (\mathbf{W} - \widehat{\mathbf{W}})\boldsymbol{\mu}\mathbf{1}^T)\right\|_F^2 \\
    &= \left\|\mathbf{WX} - \widehat{\mathbf{W}}\mathbf{X} - (\mathbf{W} - \widehat{\mathbf{W}})\boldsymbol{\mu}\mathbf{1}^T\right\|_F^2 \\
    &= \left\|(\mathbf{W} - \widehat{\mathbf{W}})\mathbf{X} - (\mathbf{W} - \widehat{\mathbf{W}})\boldsymbol{\mu}\mathbf{1}^T\right\|_F^2  \\
    &= \left\|(\mathbf{W} - \widehat{\mathbf{W}})(\mathbf{X} - \boldsymbol{\mu}\mathbf{1}^T)\right\|_F^2 \\
    &=  \left\|\Delta \mathbf{W}\widetilde{\mathbf{X}}\right\|_F^2
    \end{aligned}
\end{equation}
where $\widetilde{\mathbf{X}} = \mathbf{X} - \boldsymbol{\mu}\mathbf{1}^T$ represents the mean-centered activations. Hence, the joint objective formula simplifies entirely to:
\begin{equation}
\arg\min_{\Delta\mathbf{W}} \left\|\Delta \mathbf{W}\widetilde{\mathbf{X}}\right\|_F^2.
\end{equation}

Because the Frobenius norm separates across matrix rows, the second derivative of the simplified objective with respect to the full matrix $\Delta\mathbf{W}$ produces a block-diagonal Hessian:
\begin{equation}
\mathbf{H}_{\text{full}} = \mathbf{I}_{d_{\text{out}}}\otimes 2\widetilde{\mathbf{X}}\widetilde{\mathbf{X}}^T = \mathbf{I}_{d{\text{out}}}\otimes\mathbf{H},
\end{equation}
where $\mathbf{H} = 2\widetilde{\mathbf{X}}\widetilde{\mathbf{X}}^T$ is the shared row-wise Hessian. The inverse of a block-diagonal matrix is obtained by inverting each block:
\begin{equation}
\mathbf{H}_{\text{full}}^{-1} = \mathbf{I}_{d_{\text{out}}}\otimes\mathbf{H}^{-1}.
\end{equation}
Thus, each row update depends only on the same $d_{\text{in}}\times d_{\text{in}}$ matrix $\mathbf{H}^{-1}$.

Substituting the unscaled covariance of the centered activations, $\mathbf{C} = \frac{1}{N}\widetilde{\mathbf{X}}\widetilde{\mathbf{X}}^T$, gives
\begin{equation}
\mathbf{H} = 2\widetilde{\mathbf{X}}\widetilde{\mathbf{X}}^T = 2N\mathbf{C}.
\end{equation}
In the algorithm, the constant factor $2N$ cancels during the inverse update, allowing the exact row-wise Hessian to be replaced with the computationally efficient online covariance matrix $\mathbf{C}$.

\subsection{Additional Experiments}

\subsubsection{Experimental Setting}
All results rely on ImageNet-1K~\cite{deng2009imagenet} training data for calibration, except for the LLM experiments in Table~\ref{tab:sup_table7}, which use Wikitext-2~\cite{merity2017wiki2}. Unless otherwise specified, the calibration set size is 8192. All experiments were conducted on a single NVIDIA GeForce RTX 3090 GPU.

\subsubsection{Full Experimental Results}
This section details the complete experimental results referenced in the main paper. Table~\ref{tab:sup_table1} present the comprehensive tabular data for the calibration sweeps. Table~\ref{tab:sup_table2} details full results of layerwise pruning ratio after our framework is applied. Table~\ref{tab:sup_table3} provides whole eigenvector pruning ratio.

\begin{figure}[t]
\centering

% --- FIRST ROW ---
% --- Top Left Figure ---
\begin{minipage}{0.48\textwidth}
  \centering
  \includegraphics[width=\linewidth]{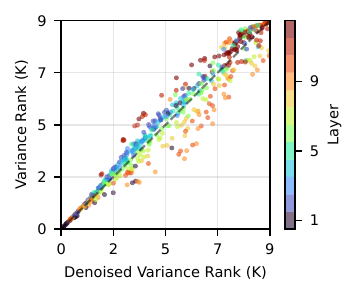}
\end{minipage}\hfill
% --- Top Right Figure ---
\begin{minipage}{0.48\textwidth}
  \centering
  \includegraphics[width=\linewidth]{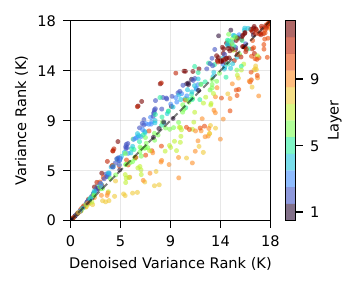}
\end{minipage}

\vspace{0.3cm} % Adds a little vertical breathing room between rows

% --- SECOND ROW ---
% --- Bottom Left Figure ---
\begin{minipage}{0.48\textwidth}
  \centering
  \includegraphics[width=\linewidth]{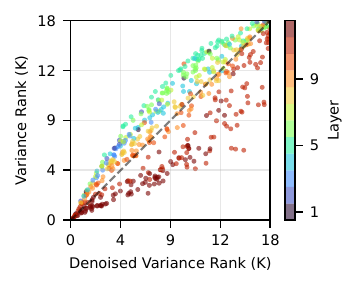}
\end{minipage}\hfill
% --- Bottom Right Figure ---
\begin{minipage}{0.48\textwidth}
  \centering
  \includegraphics[width=\linewidth]{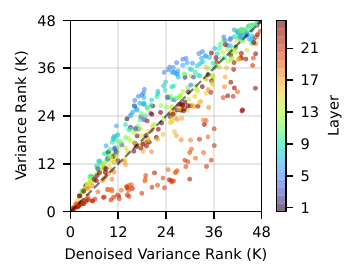}
\end{minipage}

\caption{Spearman correlation plot of DeiT-T (top left), DeiT-S (top right), Swin-T (bottom left), Swin-B (bottom right) between denoised and raw variance score rank.}
\label{fig:figure8}
\end{figure}

\begin{figure}[t]
\centering

% ==================== ROW 1 ====================
\begin{minipage}[b]{0.23\textwidth}
  \centering
  \includegraphics[width=\linewidth]{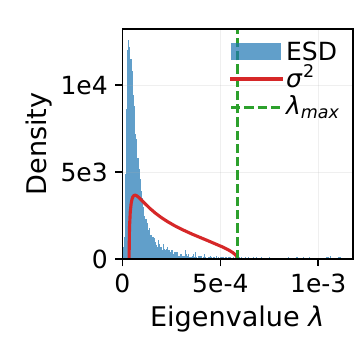}
  \centerline{(a) DeiT-B Layer 1}
\end{minipage}\hfill
\begin{minipage}[b]{0.23\textwidth}
  \centering
  \includegraphics[width=\linewidth]{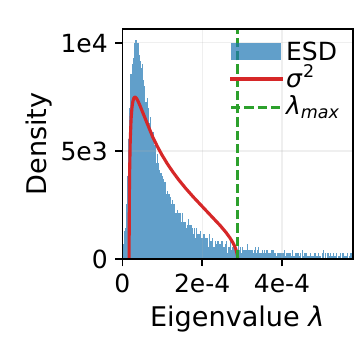}
  \centerline{(b) DeiT-B Layer 2}
\end{minipage}\hfill
\begin{minipage}[b]{0.23\textwidth}
  \centering
  \includegraphics[width=\linewidth]{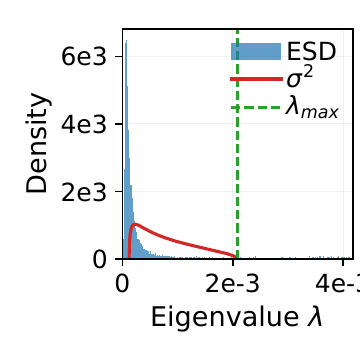}
  \centerline{(c) DeiT-B Layer 3}
\end{minipage}\hfill
\begin{minipage}[b]{0.23\textwidth}
  \centering
  \includegraphics[width=\linewidth]{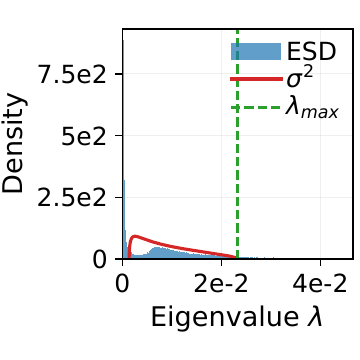}
  \centerline{(d) DeiT-B Layer 4}
\end{minipage}

\vspace{0.4cm} % <--- Blank line above this is CRITICAL to force the new row

% ==================== ROW 2 ====================
\begin{minipage}[b]{0.23\textwidth}
  \centering
  \includegraphics[width=\linewidth]{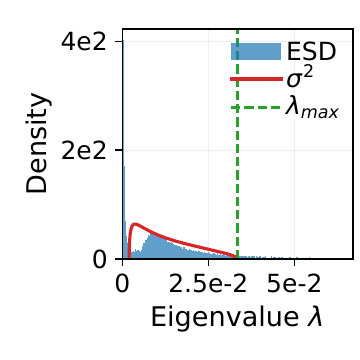}
  \centerline{(e) DeiT-B Layer 5}
\end{minipage}\hfill
\begin{minipage}[b]{0.23\textwidth}
  \centering
  \includegraphics[width=\linewidth]{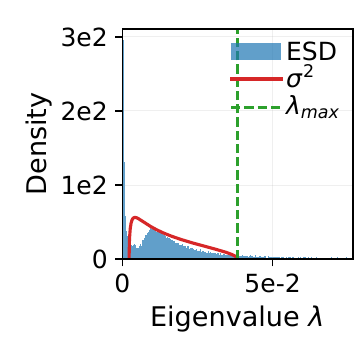}
  \centerline{(f) DeiT-B Layer 6}
\end{minipage}\hfill
\begin{minipage}[b]{0.23\textwidth}
  \centering
  \includegraphics[width=\linewidth]{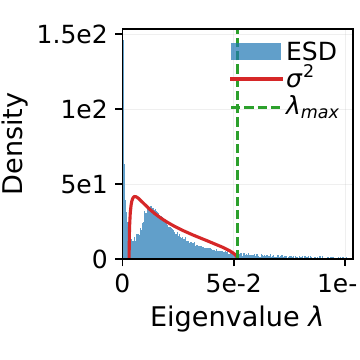}
  \centerline{(g) DeiT-B Layer 7}
\end{minipage}\hfill
\begin{minipage}[b]{0.23\textwidth}
  \centering
  \includegraphics[width=\linewidth]{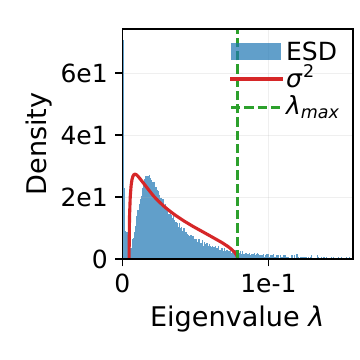}
  \centerline{(h) DeiT-B Layer 8}
\end{minipage}

\vspace{0.4cm}

% ==================== ROW 3 ====================
\begin{minipage}[b]{0.23\textwidth}
  \centering
  \includegraphics[width=\linewidth]{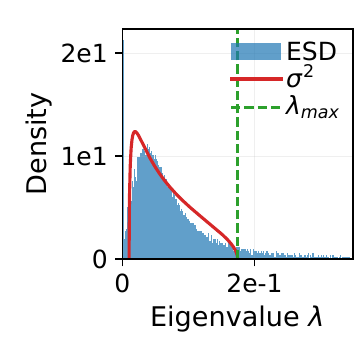}
  \centerline{(i) DeiT-B Layer 9}
\end{minipage}\hfill
\begin{minipage}[b]{0.23\textwidth}
  \centering
  \includegraphics[width=\linewidth]{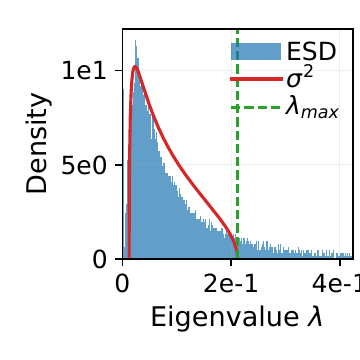}
  \centerline{(j) DeiT-B Layer 10}
\end{minipage}\hfill
\begin{minipage}[b]{0.23\textwidth}
  \centering
  \includegraphics[width=\linewidth]{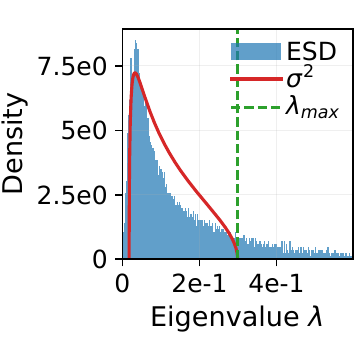}
  \centerline{(k) DeiT-B Layer 11}
\end{minipage}\hfill
\begin{minipage}[b]{0.23\textwidth}
  \centering
  \includegraphics[width=\linewidth]{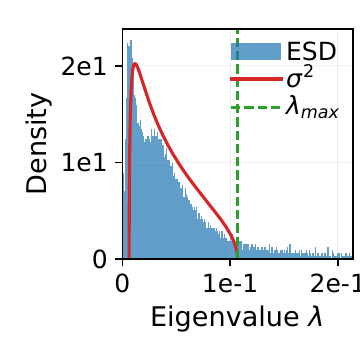}
  \centerline{(l) DeiT-B Layer 12}
\end{minipage}

\caption{Empirical spectral density for all layers of DeiT-B fitted with the Marchenko-Pastur distribution.}
\label{fig:esd_deit_b}
\end{figure}

\begin{figure}[t]
\centering

% ==================== ROW 1 ====================
\begin{minipage}[b]{0.23\textwidth}
  \centering
  \includegraphics[width=\linewidth]{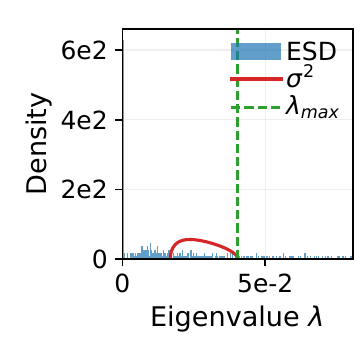}
  \centerline{(a) Swin-S Layer 1}
\end{minipage}\hfill
\begin{minipage}[b]{0.23\textwidth}
  \centering
  \includegraphics[width=\linewidth]{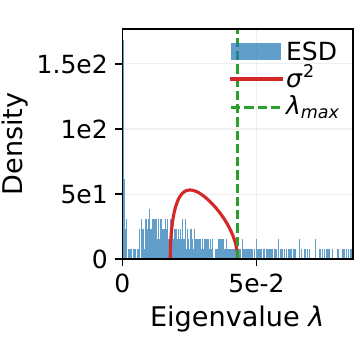}
  \centerline{(b) Swin-S Layer 2}
\end{minipage}\hfill
\begin{minipage}[b]{0.23\textwidth}
  \centering
  \includegraphics[width=\linewidth]{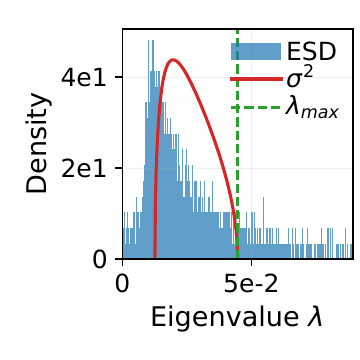}
  \centerline{(c) Swin-S Layer 3}
\end{minipage}\hfill
\begin{minipage}[b]{0.23\textwidth}
  \centering
  \includegraphics[width=\linewidth]{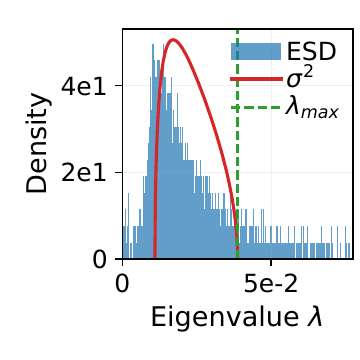}
  \centerline{(d) Swin-S Layer 4}
\end{minipage}

\vspace{0.4cm} % <--- Blank line above this is CRITICAL to force the new row

% ==================== ROW 2 ====================
\begin{minipage}[b]{0.23\textwidth}
  \centering
  \includegraphics[width=\linewidth]{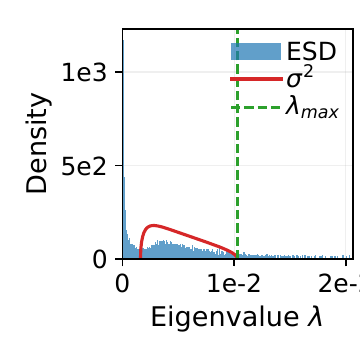}
  \centerline{(e) Swin-S Layer 5}
\end{minipage}\hfill
\begin{minipage}[b]{0.23\textwidth}
  \centering
  \includegraphics[width=\linewidth]{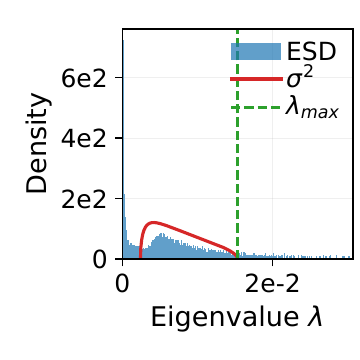}
  \centerline{(f) Swin-S Layer 6}
\end{minipage}\hfill
\begin{minipage}[b]{0.23\textwidth}
  \centering
  \includegraphics[width=\linewidth]{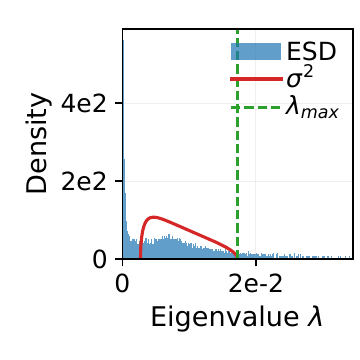}
  \centerline{(g) Swin-S Layer 7}
\end{minipage}\hfill
\begin{minipage}[b]{0.23\textwidth}
  \centering
  \includegraphics[width=\linewidth]{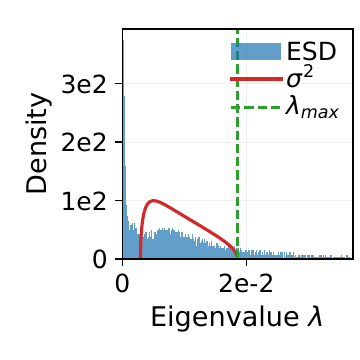}
  \centerline{(h) Swin-S Layer 8}
\end{minipage}

\vspace{0.4cm}

% ==================== ROW 3 ====================
\begin{minipage}[b]{0.23\textwidth}
  \centering
  \includegraphics[width=\linewidth]{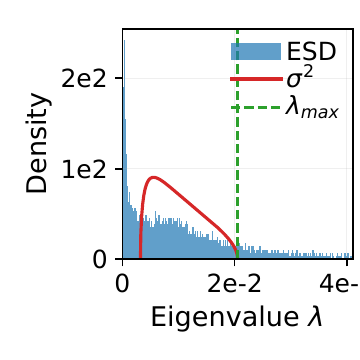}
  \centerline{(i) Swin-S Layer 9}
\end{minipage}\hfill
\begin{minipage}[b]{0.23\textwidth}
  \centering
  \includegraphics[width=\linewidth]{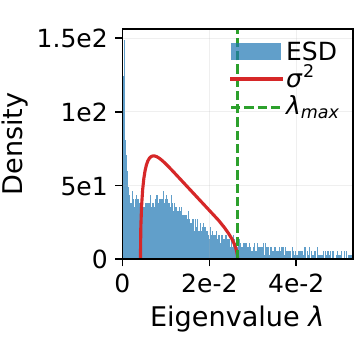}
  \centerline{(j) Swin-S Layer 10}
\end{minipage}\hfill
\begin{minipage}[b]{0.23\textwidth}
  \centering
  \includegraphics[width=\linewidth]{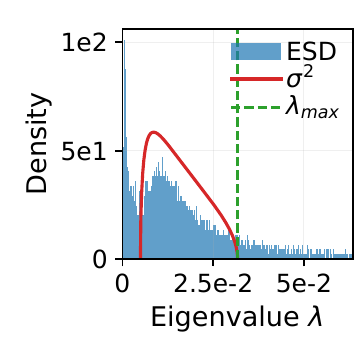}
  \centerline{(k) Swin-S Layer 11}
\end{minipage}\hfill
\begin{minipage}[b]{0.23\textwidth}
  \centering
  \includegraphics[width=\linewidth]{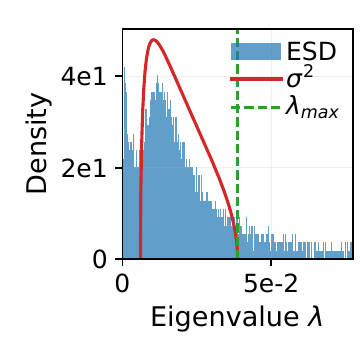}
  \centerline{(l) Swin-S Layer 12}
\end{minipage}

\caption{Empirical spectral density for layers 1-12 of Swin-S fitted with the Marchenko-Pastur distribution.}
\label{fig:esd_swin_s_1_12}
\end{figure}

\begin{figure}[t]
\centering

% ==================== ROW 1 ====================
\begin{minipage}[b]{0.23\textwidth}
  \centering
  \includegraphics[width=\linewidth]{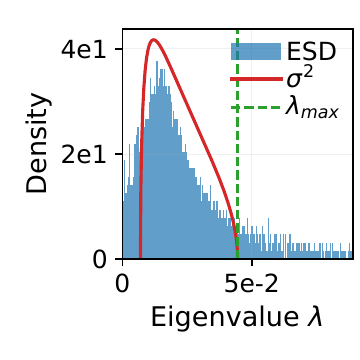}
  \centerline{(m) Swin-S Layer 13}
\end{minipage}\hfill
\begin{minipage}[b]{0.23\textwidth}
  \centering
  \includegraphics[width=\linewidth]{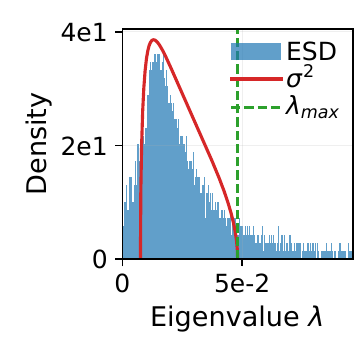}
  \centerline{(n) Swin-S Layer 14}
\end{minipage}\hfill
\begin{minipage}[b]{0.23\textwidth}
  \centering
  \includegraphics[width=\linewidth]{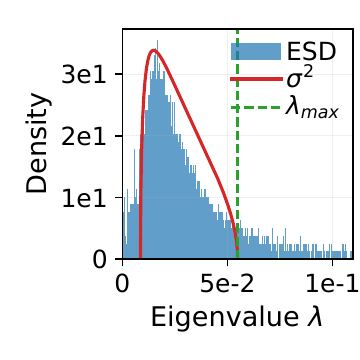}
  \centerline{(o) Swin-S Layer 15}
\end{minipage}\hfill
\begin{minipage}[b]{0.23\textwidth}
  \centering
  \includegraphics[width=\linewidth]{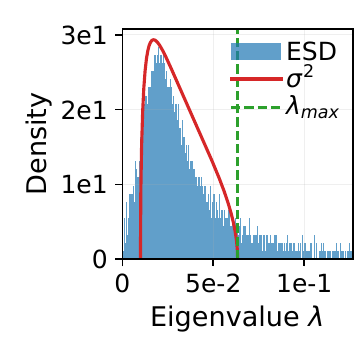}
  \centerline{(p) Swin-S Layer 16}
\end{minipage}

\vspace{0.4cm} % <--- Blank line above this is CRITICAL to force the new row

% ==================== ROW 2 ====================
\begin{minipage}[b]{0.23\textwidth}
  \centering
  \includegraphics[width=\linewidth]{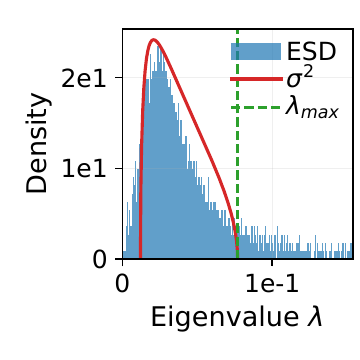}
  \centerline{(q) Swin-S Layer 17}
\end{minipage}\hfill
\begin{minipage}[b]{0.23\textwidth}
  \centering
  \includegraphics[width=\linewidth]{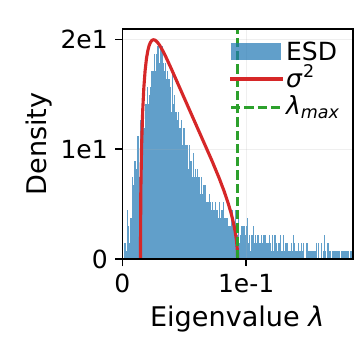}
  \centerline{(r) Swin-S Layer 18}
\end{minipage}\hfill
\begin{minipage}[b]{0.23\textwidth}
  \centering
  \includegraphics[width=\linewidth]{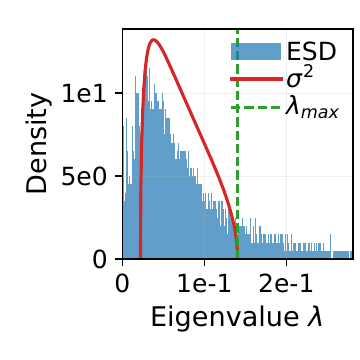}
  \centerline{(s) Swin-S Layer 19}
\end{minipage}\hfill
\begin{minipage}[b]{0.23\textwidth}
  \centering
  \includegraphics[width=\linewidth]{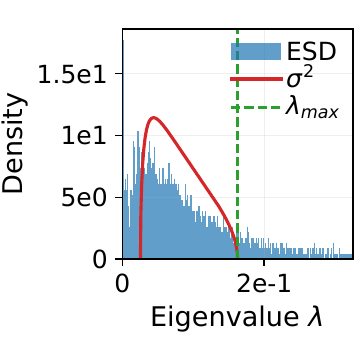}
  \centerline{(t) Swin-S Layer 20}
\end{minipage}

\vspace{0.4cm}

% ==================== ROW 3 ====================
\begin{minipage}[b]{0.23\textwidth}
  \centering
  \includegraphics[width=\linewidth]{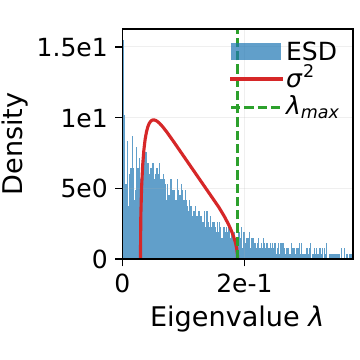}
  \centerline{(u) Swin-S Layer 21}
\end{minipage}\hfill
\begin{minipage}[b]{0.23\textwidth}
  \centering
  \includegraphics[width=\linewidth]{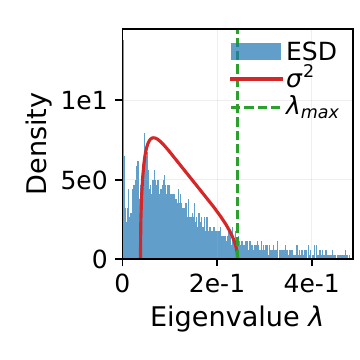}
  \centerline{(v) Swin-S Layer 22}
\end{minipage}\hfill
\begin{minipage}[b]{0.23\textwidth}
  \centering
  \includegraphics[width=\linewidth]{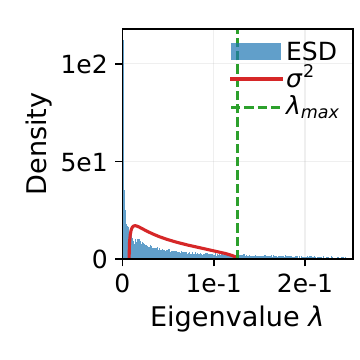}
  \centerline{(w) Swin-S Layer 23}
\end{minipage}\hfill
\begin{minipage}[b]{0.23\textwidth}
  \centering
  \includegraphics[width=\linewidth]{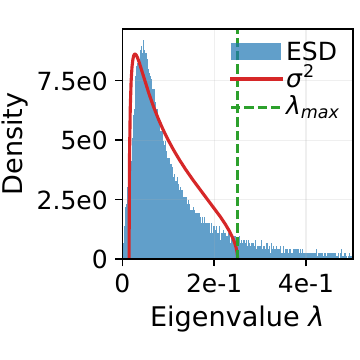}
  \centerline{(x) Swin-S Layer 24}
\end{minipage}

\caption{Empirical spectral density for layers 13-24 of Swin-S fitted with the Marchenko-Pastur distribution.}
\label{fig:esd_swin_s_13_24}
\end{figure}

Furthermore, we provide extended visualizations: Figure~\ref{fig:figure8} plots the full spearman correlation plots of denoised and raw variance scores, while Figure~\ref{fig:esd_deit_b}, Figure~\ref{fig:esd_swin_s_1_12}, and Figure~\ref{fig:esd_swin_s_13_24} illustrate the complete empirical spectral density plots for both the DeiT-B and Swin-S models.

\begin{table}[H]
\centering
\caption{Calibration set size experiments. Model pruning ratio is 50$\%$ for all models.}
\label{tab:sup_table1}
\includegraphics[width=\textwidth, keepaspectratio]{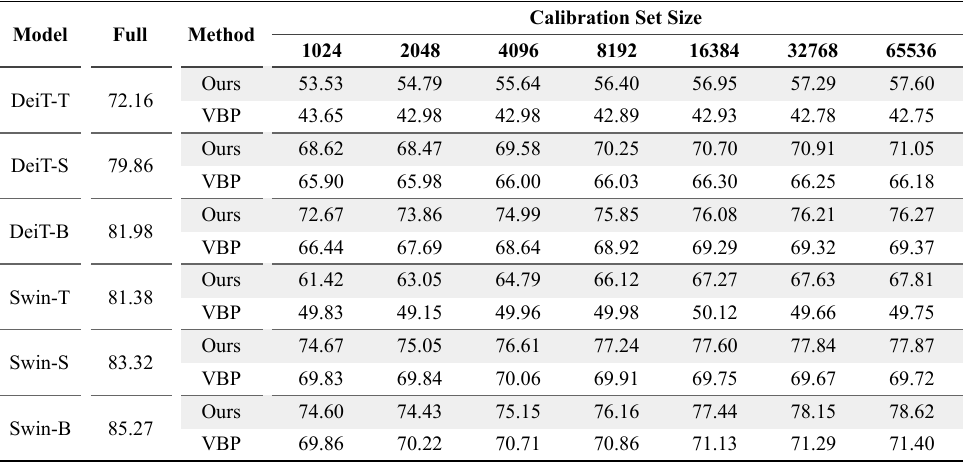}
\end{table}

\begin{table}[t]
\centering
\caption{Layerwise pruning ratios. Model pruning ratio is 50$\%$ for all models.}
\label{tab:sup_table2}
\includegraphics[width=1\textwidth, keepaspectratio]{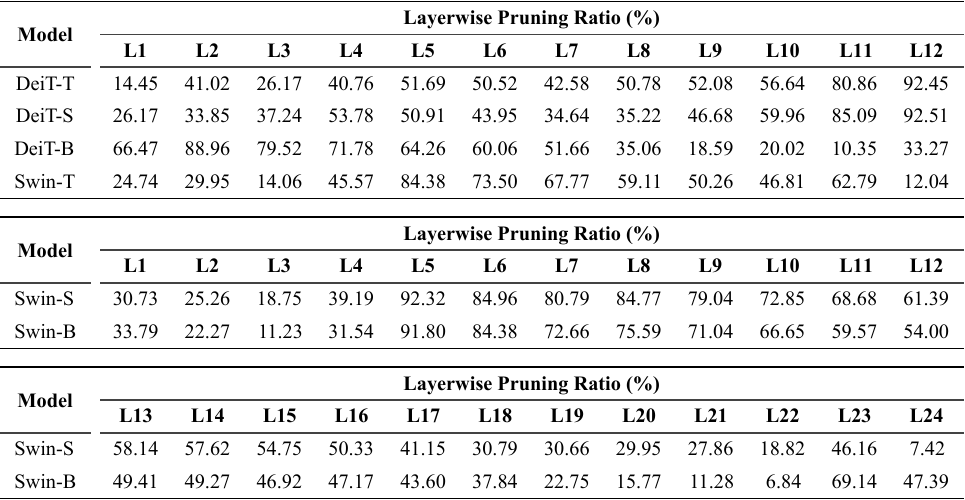}
\end{table}

\begin{table}[H]
\centering
\caption{Layerwise eigenvector pruning ratios based on Marchenko-Pastur denoising.}
\label{tab:sup_table3}
\includegraphics[width=1\textwidth, keepaspectratio]{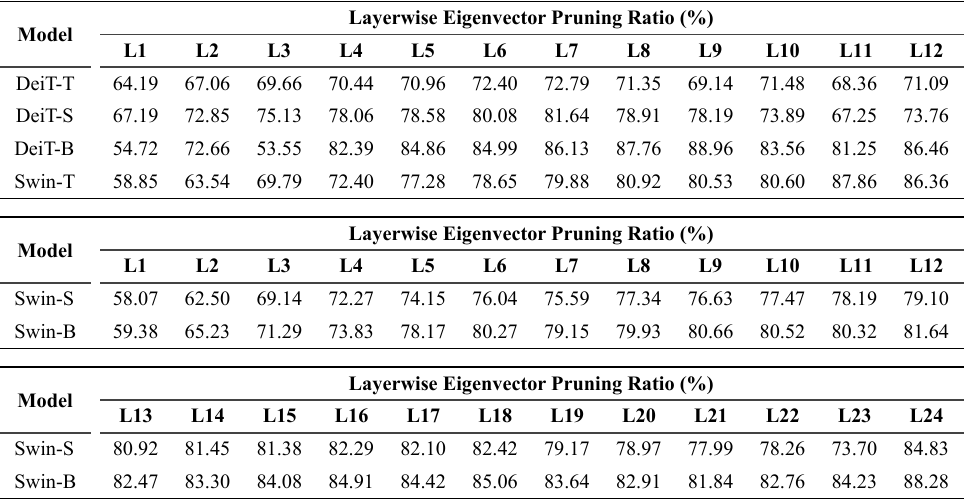}
\end{table}

\subsubsection{Further Analysis}
To rigorously validate our framework, we perform additional analysis and ablation studies. Table~\ref{tab:sup_tables4} provides comparison with additional baselines adapted for structured MLP pruning: Hessian-based NViT~\cite{Yang2023NViT}, gradient-based SynFlow~\cite{Tanaka2020SynFlow}, and training-based ViT-Slim~\cite{havan2022vision}. Our method outperforms all baselines.

Table~\ref{tab:sup_tables5} ablates the individual contributions of denoising, mean-shift, and OB$^2$C. Denoising results in better accuracy at high pruning ratio, while mean-shift and OB$^2$C increase accuracy across all pruning ratios. Table~\ref{tab:sup_tables6} demonstrates robustness to the calibration set distribution by measuring the standard deviation of Top-1 accuracy across three calibration set seeds. We see no meaningful deviation.

\begin{table}[t]
\centering
\caption{Experiment with additional baselines. Model pruning ratio is 50$\%$ for all models.}
\label{tab:sup_tables4}
\includegraphics[width=1\textwidth, keepaspectratio]{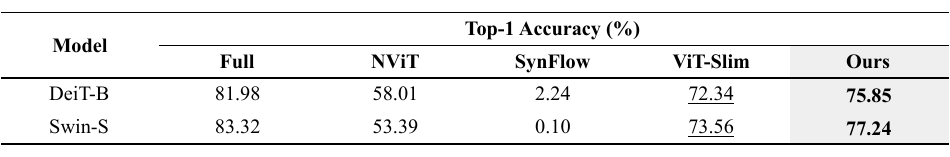}
\end{table}

\begin{table}[t]
\centering
\caption{Ablation study: denoising, mean-shift, and OB$^2$C.}
\label{tab:sup_tables5}
\includegraphics[width=1\textwidth, keepaspectratio]{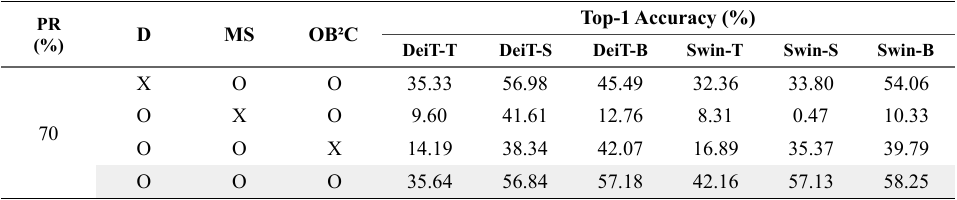}
\end{table}

\begin{table}[t]
\centering
\caption{Top-1 accuracy standard deviation measured across three calibration sets, with random seeds 100, 200, and 300.}
\label{tab:sup_tables6}
\includegraphics[width=1\textwidth, keepaspectratio]{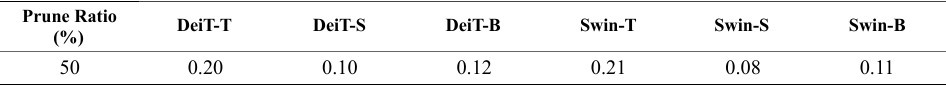}
\end{table}

\subsubsection{Scalability and Applicability}
To validate scalability, Table~\ref{tab:sup_table7} evaluates our framework on larger architectures, including ViT-L~\cite{radford2021learning} and Qwen2.5 1.5B~\cite{qwen2025qwen25technicalreport}. For Qwen's SwiGLU MLPs, we calibrate on 128 Wikitext-2 samples (2048-token length) and apply uniform layer-wise pruning based on our hybrid scores. Specifically, we prune the lowest-scoring down-projection input channels and their corresponding up and gate-projection output channels, followed by OB$^2$C updates and mean-shift compensation on the down-projection layer. On the Wikitext-2 test set, our method outperforms all baselines, including the LLM-specific LLM-Pruner~\cite{ma2023llmpruner}.

Furthermore, Table~\ref{tab:sup_tables8} demonstrates our method's generalizability to Multi-Head Attention (MHA) layers in ViT models. By averaging our hybrid scores across all attention heads, we uniformly prune the lowest-scoring QKV output channels and their corresponding out-projection input channels. With the application of OB$^2$C updates and mean-shift compensation, our approach again outperforms all evaluated baselines.

\begin{table}[t]
\centering
\caption{LLM and larger vision model pruned and calibrated on Wikitext-2 (128 samples) and ImageNet-1K (8192 samples).}
\label{tab:sup_table7}
\includegraphics[width=1\textwidth, keepaspectratio]{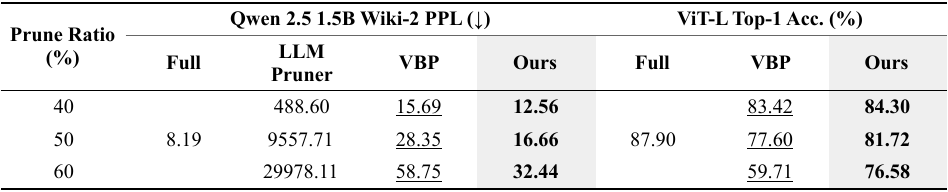}
\end{table}

\begin{table}[t]
\centering
\caption{Top-1 acc. results for structured pruning of MHA layer. Model pruning ratio is 30$\%$ for all models.}
\label{tab:sup_tables8}
\includegraphics[width=1\textwidth, keepaspectratio]{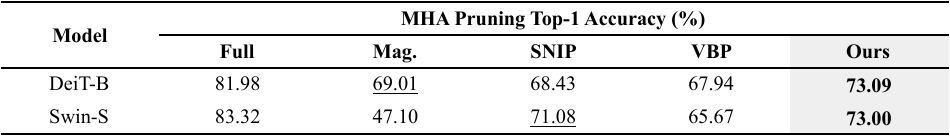}
\end{table}

